\documentclass{article}
\usepackage{amsmath}
\PassOptionsToPackage{numbers, compress}{natbib}
\usepackage[preprint]{neurips_2026}

\usepackage[utf8]{inputenc}
\usepackage[T1]{fontenc}
\usepackage{hyperref}
\usepackage{url} 
\usepackage{booktabs}
\usepackage{amsfonts}
\usepackage{nicefrac}
\usepackage{microtype}
\usepackage{xcolor}
\usepackage{wrapfig}

\usepackage{hyperref}
\usepackage{url}
\usepackage{booktabs}
\usepackage{amsfonts}
\usepackage{nicefrac}
\usepackage{microtype}
\usepackage{xcolor}
\usepackage{bm}
\usepackage{amsmath}
\usepackage{amsthm}
\usepackage{graphics}
\usepackage{graphicx}
\usepackage{subfigure}
\usepackage{multirow}
\usepackage{amssymb}
\usepackage{wrapfig}
\usepackage{float}
\usepackage{makecell}
\usepackage{diagbox}
\usepackage[linesnumbered,ruled]{algorithm2e}
\usepackage{booktabs}
\usepackage{color, colortbl}
\definecolor{Gray}{gray}{0.9}
\usepackage{capt-of}
\usepackage{varwidth}
\usepackage{enumerate}
\usepackage{threeparttable}
\usepackage{bbding}
\usepackage{arydshln}

\newtheorem{definition}{Definition}
\definecolor{brilliantrose}{rgb}{1.0, 0.33, 0.64}

\definecolor{remark}{rgb}{1,.5,0} 
\definecolor{citecolor}{rgb}{0,0.443,0.737} 
\definecolor{linkcolor}{rgb}{0.956,0.298,0.235} 

\definecolor{cyan}{rgb}{0.831,0.901,0.945}
\usepackage{comment}

\usepackage[most]{tcolorbox}
\usepackage{amsmath}
\usepackage{amssymb}
\usepackage{caption}

\usepackage{titletoc}
\usepackage{xcolor}
\usepackage{hyperref}
\usepackage{titlesec}
\usepackage{lineno}
\definecolor{appendixblue}{RGB}{0,114,188}

\newcommand{\appendixtableofcontents}{

    \begingroup
    \nolinenumbers

    {\Huge\bfseries Appendix\par}
    \vspace{1.5em}
    {\Large\bfseries Table of Contents\par}
    \vspace{0.4em}
    \hrule
    \vspace{1em}

    \titlecontents{section}
      [0em]    
      {\vspace{0.6em}\large\bfseries\color{appendixblue}}
      {\contentslabel{2.3em}}
      {}
      {\hfill\contentspage} 
      []

    \titlecontents{subsection}
      [2.3em]
      {\normalsize\color{appendixblue}}
      {\contentslabel{2.8em}}
      {}
      {\titlerule*[0.6pc]{.}\contentspage}
      []

    \printcontents[appendix]{}{1}{\setcounter{tocdepth}{2}}

    \par\vspace{0.8em}
    \hrule
    \endgroup
}

\colorlet{dark-blue}{blue!65!black}
\colorlet{dark-green}{green!55!black}
\colorlet{dark-red}{red!80!black}
\colorlet{dark-yellow}{yellow!90!black}
\colorlet{white-blue}{blue!70!green}
\definecolor{mypink}{RGB}{219, 48, 122}
\hypersetup{
    colorlinks=true,%
    citecolor=gray,%
    filecolor=citecolor,%
    linkcolor=citecolor,%
    urlcolor=magenta
}
\newcommand{\ourdata}{UMMSteer\xspace}
\usepackage{enumitem}

\title{Do Unified Multimodal Models Think in One Space?\\
A Lens Through Cross-Branch Steering
}

\author{
 \textbf{Yu Wang},
 \textbf{Sharon Li}
\\
Department of Computer Sciences, University of Wisconsin-Madison
\\
  \texttt{\{yuwang, sharonli\}}@cs.wisc.edu
}

\begin{document}

\maketitle
\begin{abstract}
Unified multimodal models (UMMs) aim to integrate understanding and generation within a single architecture, yet it remains unclear whether these capabilities share a unified and transferable semantic space.
This question is fundamentally challenging, as the two branches operate over heterogeneous representations (text tokens vs.\ visual latents) and distinct training objectives, making direct comparison difficult. 
To address this, we introduce \emph{cross-branch semantic steering}, an intervention-based framework that extracts semantic directions from one branch and applies them to the other. 
We show that steering vectors learned from the understanding branch can transfer to generation, enabling controllable image synthesis and improved semantic faithfulness. In contrast, the reverse direction consistently shows limited effectiveness. Our analysis suggests that this asymmetry may be related to a practical representational mismatch: understanding-derived vectors capture transferable, object-centric semantics, while generation-derived vectors primarily encode low-level appearance features.
Our results reveal that architectural unification does not guarantee semantic alignment, and establish cross-branch steering as a practical tool for probing multimodal representations.
\end{abstract}

\section{Introduction}
\label{sec:intro}
Unified Multimodal Models (UMMs) integrate multimodal understanding and generation within a single architecture, and have emerged as a pivotal step toward artificial general intelligence~\cite{bagel,januspro,qwenimage,unipic,unimodel,uniworld,showo2,zhao2026unifiedmultimodalunderstandinggeneration,qu2025tokenflow,xie2025reconstructionalignmentimprovesunified,wu2025harmonizingvisualrepresentationsunified,wu2025openunisimplebaselineunified,luo2026torchummunifiedmultimodalmodel,su2026unigameturningunifiedmultimodal,yan2026unifiedmultimodalmodelsautoencoders,sun2026rethinkingummvisualgeneration,liu2025tunatamingunifiedvisual,wu2026omnigen2instructionalignedmultimodalgeneration,jiao2025unitoken,zheng2025architecture}. These models maintain two functional branches through a shared backbone: an \emph{understanding branch} that encodes visual and textual inputs into semantic representations for tasks such as visual question answering and captioning, and a \emph{generation branch} that synthesizes images from latent representations conditioned on text. While recent UMMs have demonstrated strong performance across both understanding and generation tasks~\cite{bagel,januspro,qwenimage,showo2,luo2026torchummunifiedmultimodalmodel,su2026unigameturningunifiedmultimodal}, a fundamental question remains open: \emph{Do the unified architectures of UMMs give rise to unified representations,  and do the understanding and generation branches share a common semantic space?}
\begin{figure}[t]
\centering
\includegraphics[width=\linewidth]{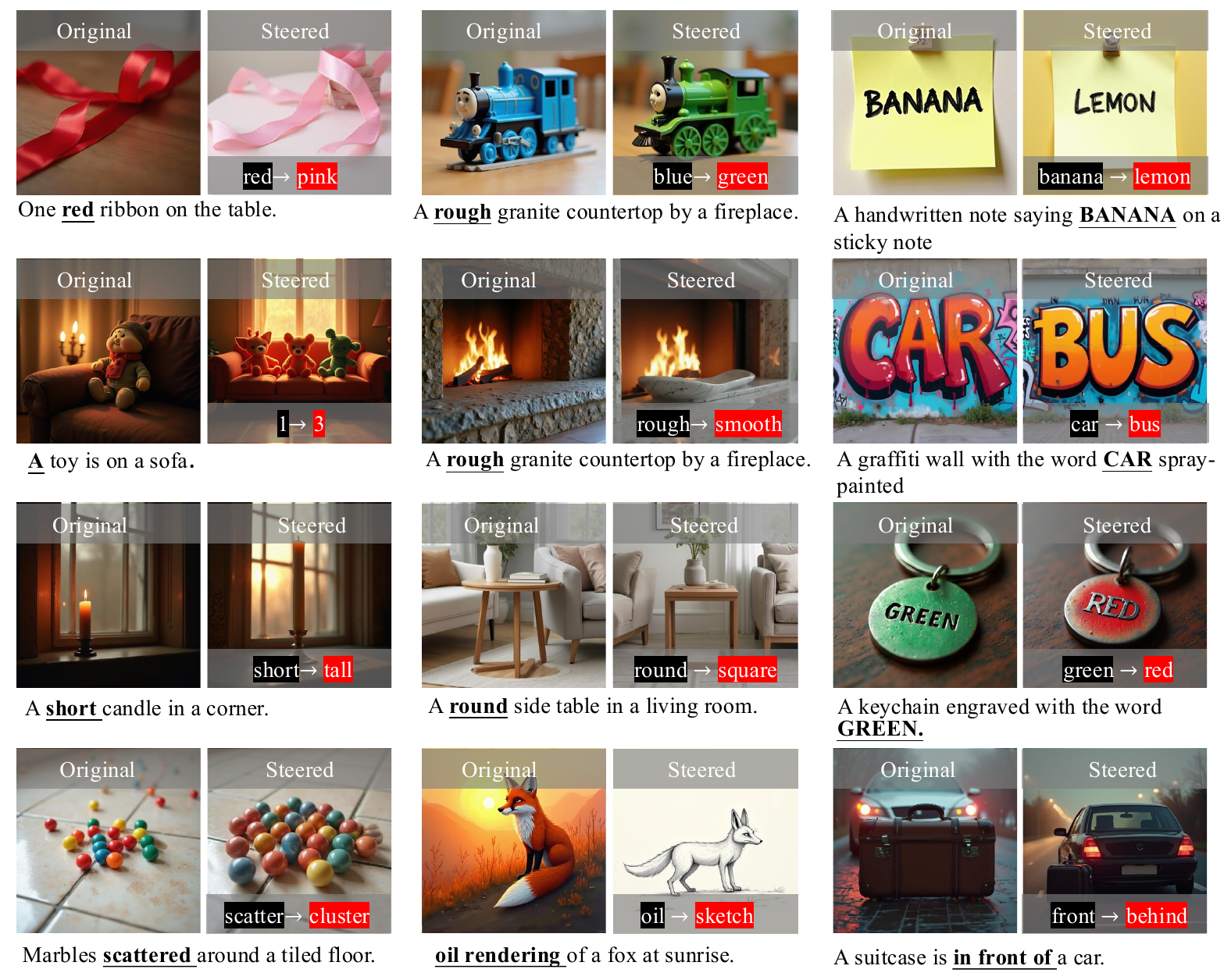}
\caption{
{Qualitative results of understanding $\rightarrow$ generation steering} across  semantic types. Each pair shows the original generation (left) and the steered output (right). Steering modifies the target attribute while preserving object identity. More results are provided in the appendix~\ref{asec:visual_case}.
}
\label{fig:intro}
\end{figure}

A unified semantic space would imply that knowledge acquired through understanding about objects, attributes, and relations is directly accessible to generation, enabling tighter controllability and improved faithfulness~\cite{yan2026unifiedmultimodalmodelsautoencoders,wang2026quantifying,li2025unievalunifiedholisticevaluation}. A fragmented space, by contrast, would mean that current UMMs are architecturally unified but representationally disjoint, a structural limitation that standard benchmarks cannot detect. Despite the importance, answering this question is non-trivial: the two branches operate on heterogeneous token types (text tokens vs.\ visual latents), optimize for different objectives; therefore, whether semantic structure can transfer across this boundary remains unknown.

In this paper, we introduce {cross-branch semantic steering} as a diagnostic framework to probe this question. Rather than comparing representations directly, we employ an intervention-based test: extract a steering vector encoding a target semantic direction (e.g., ``red $\rightarrow$ pink'') from one branch and inject it into the other during inference. If a vector learned from the understanding branch causes the generation branch to produce a pink object instead of a red one, this constitutes direct, causal evidence of a unified semantic space. While steering has been widely used to probe LLM internals~\cite{caa, pca, cls}, prior work operates within a single modality. Whether steering vectors can transfer \emph{across} modality branches---where token types, training signals, and representational structures differ---is an open question. To our knowledge, this is the first work to explore the possibility and impossibility of steering across understanding and generation branches in UMMs. 

In this paper, we show that steering vectors learned from the understanding branch can transfer to the generation branch. To operationalize this, we construct UMMSteer, a visually grounded contrastive dataset comprising 51 diverse concept pairs spanning {color}, {text}, {counting}, {position}, {style}, {appearance}, and {persona}. For each concept pair, we first extract a semantic steering vector from the understanding branch using text-based QA samples, and then apply this vector to the generation branch to test whether it can causally shift image generation. 
As illustrated in Fig.~\ref{fig:intro}, using a vector encoding ``red $\rightarrow$ pink'', The model produces a pink ribbon when prompted with ``One red ribbon on the table.''
Notably, these vectors are learned entirely from textual representations, yet their effect can manifest and transfer visually. In other words, the understanding branch's semantic knowledge directly reshapes what the generation branch produces. 

Meanwhile, we uncover a striking asymmetry: while understanding-to-generation transfer is feasible, the reverse direction consistently fails. Steering vectors derived from the generation branch do not meaningfully influence the understanding branch's predictions.
Our analysis suggests that this asymmetry may stem from a representational mismatch: understanding-derived vectors tend to capture compositional, object-centric semantics that align well with both branches, while generation-derived vectors primarily capture low-level appearance features---background tone, illumination, global color statistics---that show weaker alignment with the understanding branch's semantic structure. 
Cross-branch transfer succeeds only when the steering direction resides within a shared subspace defined by understanding-level abstractions. To verify the generalizability of our findings, we evaluate cross-branch steering on other state-of-the-art UMMs, including Janus-Pro~\cite{januspro} and UniPic-1~\cite{unipic}. Our analysis suggests that strong model performance does not necessarily imply aligned semantic representations across branches. Instead, such alignment depends on the underlying architectural design and modality integration.
We summarize our main contributions as follows:
\vspace{-5pt}
\begin{enumerate}
    \item We introduce \emph{cross-branch semantic steering}, the first framework to transfer steering vectors across understanding and generation branches in UMMs. We construct a visually grounded contrastive steering dataset for this purpose, which will be publicly released.
    \item We uncover a \emph{fundamental asymmetry in cross-branch transfer}: semantic directions learned from the understanding branch can control generation, while the reverse direction consistently fails. 
    \item We trace this asymmetry to a \emph{representational mismatch between branches}, pointing toward concrete design principles for building UMMs with genuinely unified semantic spaces.
\end{enumerate}
\section{Related Work}
\paragraph{Unified Multimodal Models.}
Unified multimodal models (UMMs) aim to build a single architecture capable of both understanding and generating data across multiple modalities~\cite{uniworld,li2025unifork,qin2026unicotunifiedchainofthoughtreasoning,xiao2025mindomniunleashingreasoninggeneration,chen2025blip3ofamilyfullyopen,tong2024metamorphmultimodalunderstandinggeneration,zhou2025hermesunifiedselfdrivingworld,sun2024generativemultimodalmodelsincontext}. These models typically maintain two functional branches: an \emph{understanding branch} that encodes visual and textual inputs into semantic representations for tasks such as visual question answering and captioning, and a \emph{generation branch} that synthesizes images from latent representations conditioned on text. Although the two branches differ in their input encoders and output mechanisms, they typically share a common backbone.
Existing UMMs fall into three broad paradigms. Pure autoregressive models serialize visual and textual tokens into a unified sequence and model them via next-token prediction, as in Janus-Pro~\cite{januspro} and UniPic-1~\cite{unipic}. Pure diffusion-based approaches perform both understanding and generation within a unified diffusion framework, as in UniModel~\cite{unimodel}. Hybrid models combine autoregressive text generation with diffusion- or flow-based image synthesis, using language modeling for semantic control and diffusion for high-fidelity visual generation, as in  BAGEL~\cite{bagel} and Show-o2~\cite{showo2}. While these models demonstrate strong performance on standard benchmarks, it remains unclear whether their unified architectures give rise to unified representations,  and whether the understanding and generation branches share a common semantic space. Our work directly probes this important question through the lens of cross-branch steering. 

\paragraph{Steering and Representation Intervention in LLMs.} 
Steering methods modify model behavior at inference time by adding learned directions to hidden states. A central family of approaches derives steering vectors from contrastive representations: mean differences (CAA)~\cite{caa}, principal components of difference vectors~\cite{pca}, linear classifiers trained on hidden states~\cite{cls}, and paired contrastive prompts~\cite{turner2024steeringlanguagemodelsactivation}. Extensions include input-adaptive scaling~\cite{lee2025programmingrefusalconditionalactivation} and layer-specific multi-behavior control~\cite{vanderweij2024extendingactivationsteeringbroad}. A broader line of work on representation probing and intervention~\cite{xu2025easysteer, wang2025enhancing, zhao2026understanding, fang2026controllable, sinii2025steeringllmreasoningbiasonly, zhao2025steerxdisentangledsteeringllm,park2025steeringguidancepersonalizedtexttoimage,patel2024flowchef,li2026steeringlargereasoningmodels} has established steering as a reliable tool for analyzing and controlling LLM internals. However, all existing steering work operates within a single modality---typically text-only LLMs or, more recently, multimodal language models in their understanding capacity~\cite{wang2025steering,parekh2025learning,wu2025automating,gan2025textual,liu2025steering,zhao2025sce2drivex,zhang2025evaluating,listeering}. No prior work has attempted to transfer steering vectors across understanding and generation branches in UMMs. Our work is the first to do so, repurposing steering not as a control mechanism but as a diagnostic probe for cross-modal semantic alignment in UMMs.

\section{Cross-Branch Steering as a Diagnostic Framework}
\label{sec:method}

We aim to determine whether the understanding and generation branches of UMMs share a unified semantic space. Directly comparing representations across branches is difficult, as they differ in both token types and optimization objectives. To sidestep the challenge, we propose an intervention-based perspective: if a semantic direction extracted from one branch can reliably influence the behavior of the other, this constitutes evidence that both branches encode compatible semantic structure. 

Steering provides a natural yet unexplored lens for this purpose. In LLMs, steering vectors have been shown to capture interpretable semantic directions in hidden space, enabling controlled shifts in model behavior~\cite{caa, pca, cls}. While prior steering work operates within a single modality, the cross-branch setting in UMMs poses a distinct challenge: the two branches process different token types, serve different objectives, and may organize semantics at different levels of abstraction. \emph{Whether steering vectors can cross this boundary is an open question}. 
We emphasize that steering here serves as a {diagnostic probe} for semantic alignment, not as a proposed method for controlling generation. 
 
This section describes the setup:  the contrastive dataset UMMSteer we construct (Sec.~\ref{sec:dataset}), and how steering vectors are extracted based on the dataset (Sec.~\ref{sec:steer_learn}). The results of applying this framework will be presented in Sec.~\ref{sec:results}.

\subsection{UMMSteer: Contrastive Steering Dataset}
\label{sec:dataset}

A key challenge in operationalizing cross-branch steering is the lack of suitable data. Existing steering datasets are not directly applicable, since steering in UMMs requires concepts that are both semantically meaningful for the understanding branch and visually verifiable in the generation branch. We therefore begin by constructing UMMSteer, a contrastive dataset tailored for UMMs.

\paragraph{Semantic Concept for Steering.}
To comprehensively cover the semantic space, we define {7 semantic types} spanning {3 levels} of increasing complexity, yielding \textbf{51 pairs of semantic concepts} in total. At the {object level}, we include intrinsic properties of individual objects: \texttt{color} (e.g., red vs.\ blue), \texttt{text} (e.g., ``23'' vs.\ ``10''), and \texttt{counting} (e.g., one vs.\ four). At the {relational level}, we include spatial relationships between objects: \texttt{position} (e.g., left/right, above/below, inside/outside). At the {scene level}, we include global appearance and holistic properties: \texttt{style} (e.g., realistic vs.\ cartoon), \texttt{appearance} (e.g., clean vs.\ dirty, wooden vs.\ metal), and \texttt{persona} (e.g., neurotic vs.\ calm). We summarize the complete list of semantic concepts in Appendix~\ref{asec:data} (Tab.~\ref{tab:semantic_concepts}).

\vspace{0.2cm}
\begin{definition}[\textbf{Cross-Branch Steering Dataset}]
For a target concept $k$, we construct a dataset of $N$ contrastive pairs $\mathcal{D}^k = \{(x^+_i, x^-_i)\}_{i=1}^{N}$, where $x^+_i = (q^+_i, y^+_i)$ and $x^-_i = (q^-_i, y^-_i)$ denote the positive and negative instances, respectively. Here, $q_i$ is the input context (a question for understanding, or a text prompt for generation) and $y_i$ is the expected output (an answer token or a generated image).  Each pair differs only in the target semantic attribute while all other contextual factors remain fixed.
\end{definition}

Each pair is constructed in two branch-specific formats. We provide representative examples in Appendix (Fig.~\ref{fig:sample}). Specifically, for the understanding branch, we construct contrastive question--answer pairs where the questions describe identical scenes except for the target attribute, e.g.:
\begin{align*}
    x^+_i &: \text{``A \textbf{blue} car parked on a street. What color is this car?''} \rightarrow \textbf{blue} \\
    x^-_i &: \text{``A \textbf{red} car parked on a street. What color is this car?''} \rightarrow \textbf{red}
\end{align*}
For the generation branch, we construct contrastive text prompts that differ only in the target attribute (e.g., ``A {blue} car parked on a street'' vs.\ ``A {red} car parked on a street''). For each binary concept, we construct 50 pairs per format for learning and 50 held-out samples for evaluation.

\subsection{Learning Steering Vectors for Cross-Branch Transfer}
\label{sec:steer_learn}
A steering vector represents an additive direction in the hidden-state space that can bias a model toward a target semantic concept~\cite{caa,pca,cls}. In our setting, the goal is to learn such semantic directions from one branch of a UMM and examine whether applying them to the other branch can steer its behavior toward the same target concept. Here, we briefly describe how we learn steering vectors using representative LLM steering methods and how we apply them for cross-branch transfer.

\begin{figure}
    \centering
    \includegraphics[width=0.8\linewidth]{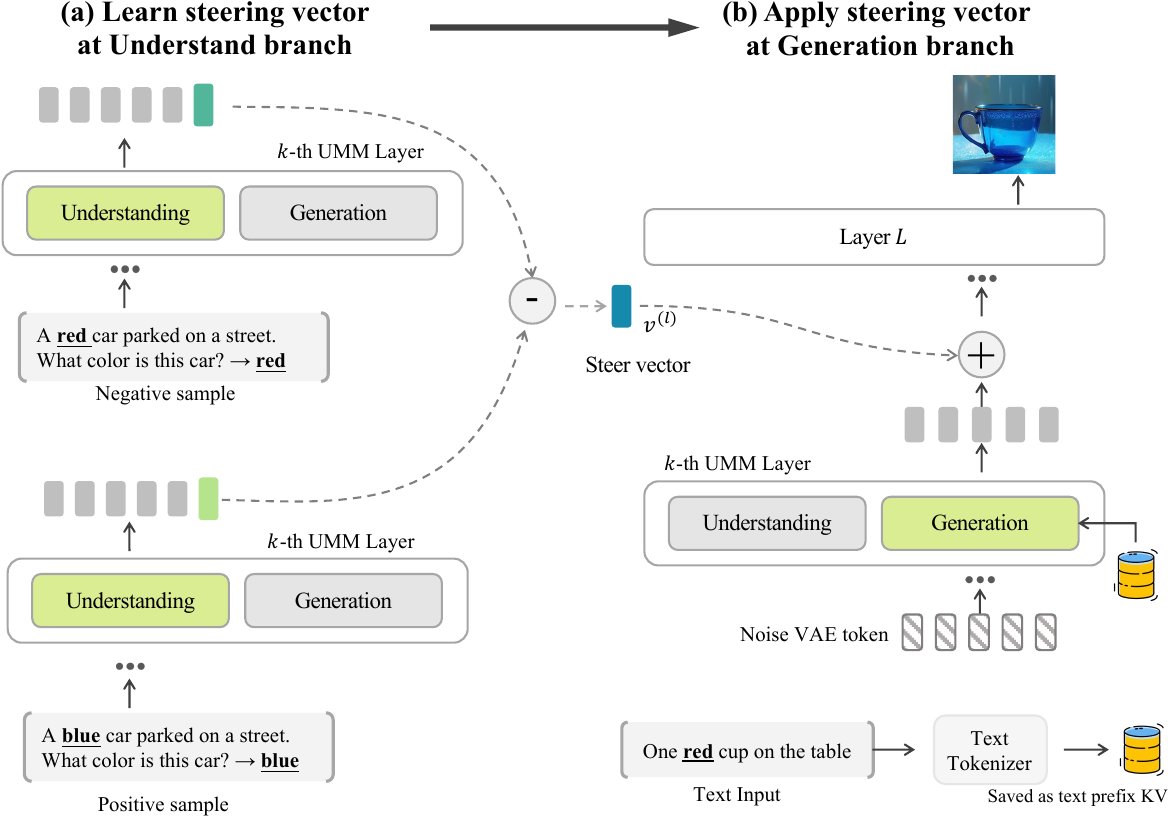}
    \vspace{-5pt}
\caption{
\textbf{Cross-branch steering pipeline (Und $\rightarrow$ Gen).}
(\textbf{a}) Contrastive text pairs (e.g., \texttt{`blue'} vs.\ \texttt{`red'}) are fed through the understanding branch. At each layer, we extract residual stream activations from the answer token and compute steering vectors.
(\textbf{b}) 
At inference time, the input text prompt (e.g., \texttt{`one red cup'}) is encoded into a prefix KV cache. During denoising, steering vectors are added to VAE token representations at each layer, biasing generation toward the target attribute.
}
\label{fig:und2gen}
\end{figure}

As shown in Fig.~\ref{fig:und2gen} (a), for a target concept $k$, we feed contrastive pairs from $\mathcal{D}^k$ into the source branch and extract residual stream activations at the answer token position (for understanding) or all VAE token positions (for generation), yielding pairs $\{\mathbf{h}^{(l)}_{+,i}, \mathbf{h}^{(l)}_{-,i}\}$. We estimate $\mathbf{v}^{(l)}$ using three common representative methods: {CAA}~\cite{caa} (mean difference between positive and negative activations), {RepE}~\cite{pca} (first principal component of the difference vectors), and {ITI}~\cite{cls} (normal vector of a learned decision boundary). All three recover a semantic direction along which model behavior can be shifted. Formal definitions are in Appendix~\ref{asec:steeringllm}.

\section{Do the Branches Share a Unified Semantic Space?}
\label{sec:results}

We now apply the cross-branch steering diagnostic framework to answer our central question. We first test whether steering vectors from the understanding branch can control generation (Sec.~\ref{sec:und_to_gen}), then examine the reverse direction (Sec.~\ref{sec:gen_to_und}), and finally analyze why transfer is asymmetric (Sec.~\ref{sec:why_asymmetry}).

\subsection{Understanding $\rightarrow$ Generation}
\label{sec:und_to_gen}

\paragraph{Applying learned steer vectors to the generation branch.}
After obtaining layer-wise steer vectors from the understanding branch, we transfer them to the {generation} branch at inference time to bias the denoising trajectory toward the desired semantic attribute, as shown in Fig~\ref{fig:und2gen}(b). Given an input prompt such as \texttt{``one red cup on the table''}, the text tokens are first encoded by the text tokenizer and processed by the model to produce a text prefix key-value cache.
This prefix KV cache serves as the textual conditioning context for image generation.
The generation process starts from noisy VAE latent tokens, which are iteratively updated through the transformer and finally decoded by the VAE decoder into an image. 

During generation, at each transformer layer $l$, the VAE latent tokens attend to the text prefix KV cache through multimodal self-attention.
We intervene on the hidden representation of the VAE tokens by adding the pre-computed steer vector $\mathbf{v}^{(l)}$:
\begin{equation}
\tilde{\mathbf{H}}_t^{(l)} = \mathbf{H}_t^{(l)} + \alpha \mathbf{v}^{(l)},
\end{equation}
where $\mathbf{H}_t^{(l)}$ denotes the hidden states of $t$-th the VAE token at layer $l$, $\mathbf{v}^{(l)}$ is the steer vector learned from the understanding branch, and $\alpha$ is the steering strength. In practice, the same steer vector is added to all VAE token representations at the selected layer, biasing the denoising trajectory toward the target semantic direction.

\paragraph{Evaluation Setup.}
To evaluate steering effectiveness in the generation branch, we construct 50 held-out text prompts for each semantic concept. The model performs text-to-image generation under steering, and the resulting images are evaluated for semantic correctness.
We adopt a question-based evaluation protocol using Qwen-VL-3 30B~\cite{qwen3vl} as the evaluator~\cite{niu2025wise,geneval}.
The evaluation consists of two stages. First, we verify whether the generated image preserves the target object, ensuring that steering does not alter object identity. Second, we assess whether the desired semantic attribute (e.g., spatial relation or color) is correctly enforced. A sample is considered successfully steered only if both conditions are satisfied.
We report the \emph{steering success rate} (SSR) as the fraction of samples that meet both criteria. More details are provided in Appendix~\ref{asec:eval_metric}. We further validate this automatic evaluation protocol through a human agreement study in Appendix~\ref{asec:human_agreement}, where Qwen-VL-3 30B achieves high agreement with human annotations, with an overall Cohen's $\kappa$ of 0.894.

\begin{table}[!t]
    \caption{\textbf{Steering success rate (SSR)  across diverse semantic categories} in BAGEL. SSR measures the fraction of generated images that both preserve the target object and correctly realize the steered attribute. Values are mean $\pm$
 std across concept pairs within each category.}
     \small
    \centering
    \begin{tabular}{p{1.5cm}<{\centering}|p{1.2cm}<{\centering}p{1.2cm}<{\centering}p{1.4cm}<{\centering}|p{1.4cm}<{\centering}|p{1.3cm}<{\centering}p{1.3cm}<{\centering}p{1.3cm}<{\centering}}
    \toprule
    & \multicolumn{3}{c|}{Object-level} & Relational & \multicolumn{3}{c}{Scene-level} \\
    \cmidrule(lr){2-4} \cmidrule(lr){5-5}  \cmidrule(lr){6-8}
    Method & Color & Text & Counting & Position & Style & Appearance & Persona \\
    \midrule
    CAA~\cite{caa} & 96.20{$_{\pm1.46}$} &\textbf{43.46{$_{\pm28.79}$}} & \textbf{68.75{$_{\pm23.96}$}} & 78.60{$_{\pm16.57}$} & \textbf{82.71{$_{\pm24.57}$}} & \textbf{79.88{$_{\pm21.06}$}} & 24.37{$_{\pm31.25}$} \\
    {RepE}~\cite{pca} & 52.95{$_{\pm40.63}$} & 21.13{$_{\pm22.50}$}  & 6.87{$_{\pm7.79}$} & 59.83{$_{\pm21.74}$} & 29.29{$_{\pm26.61}$}  & 44.25{$_{\pm23.95}$} & \textbf{34.00{$_{\pm33.94}$}}
 \\ 
  ITI~\cite{cls} & \textbf{96.48{$_{\pm1.66}$}} & 43.09{$_{\pm29.06}$} &34.90{$_{\pm28.35}$} &\textbf{78.96{$_{\pm17.32}$}}&80.75{$_{\pm21.70}$} & 78.86{$_{\pm21.80}$} & 29.38{$_{\pm30.58}$}\\
 
 \bottomrule
    \end{tabular}
    \label{tab:u2g}
\end{table}

\begin{figure}[h]
    \centering
    \includegraphics[width=0.85\linewidth]{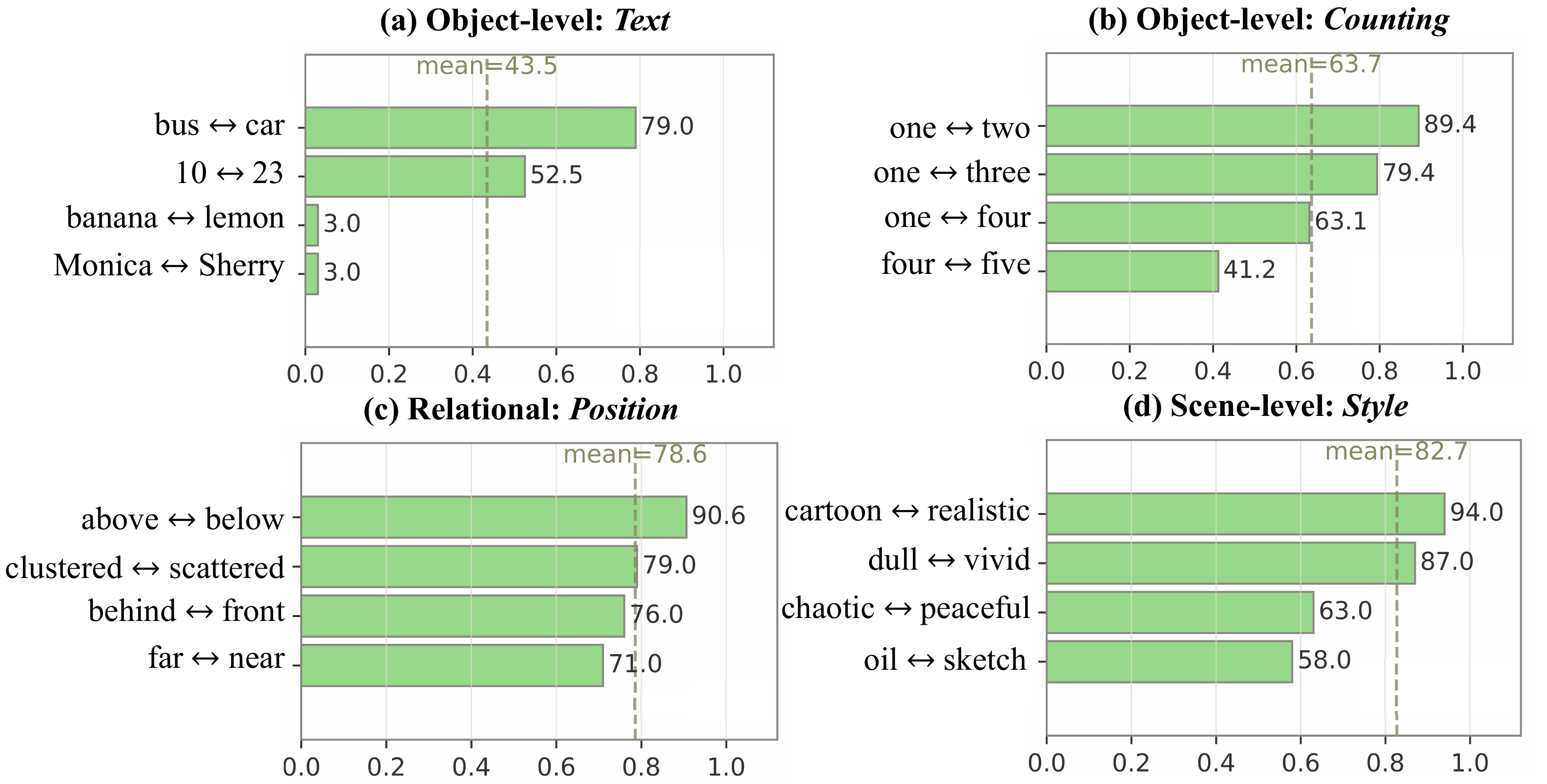}
    \vspace{-0.1cm}
\caption{\textbf{The steering success rates (SSR) across multiple semantic categories.}
The effectiveness of CAA-based steering varies within object-level, relational-level, and scene-level semantics. Full per-concept results are provided in Appendix~\ref{asec:steer_effective}.}
\label{fig:detial_ssr}
\end{figure}

\paragraph{Steering Effectiveness.}
We presents the steering success rates across all semantic categories based on BAGEL~\cite{bagel} in Tab.~\ref{tab:u2g}. Across all three steering methods, understanding-derived vectors can control the generation. However, the effectiveness is not uniform across concepts. We observe substantial within-category variance in Tab.~\ref{tab:u2g}, and therefore further present the SSR of representative concepts using CAA~\cite{caa} in Fig.~\ref{fig:detial_ssr}. 
As shown in Fig.~\ref{fig:detial_ssr}, effectiveness varies within categories and correlates with generation difficulty: simpler concepts (e.g., \texttt{one}$\rightarrow$\texttt{two} in counting category) achieve higher success rates than harder ones (e.g., \texttt{one}$\rightarrow$\texttt{four}), suggesting that steering is constrained by the model's baseline generative capability. 

Qualitative examples in Fig.~\ref{fig:intro} further illustrate successful steering across diverse concepts, including color, counting, text, appearance, and spatial relations. 
We provide additional qualitative examples in Appendix~\ref{asec:visual_case} and representative failure cases in Appendix~\ref{asec:fase_case}.

\subsection{Generation $\rightarrow$ Understanding}
\label{sec:gen_to_und}

\begin{figure}[!t]
\centering
\includegraphics[width=0.9\linewidth]{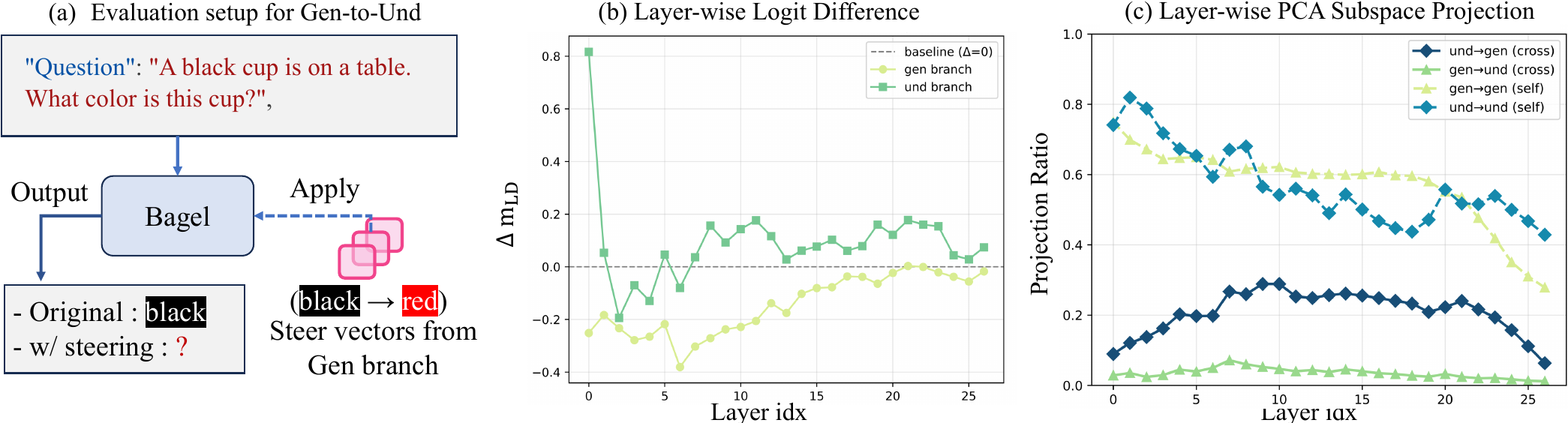}
\vspace{-5pt}
\caption{\textbf{Analysis of cross-branch (Gen → Und) steering failure.} \textbf{(a) Setup.} Steering vectors learned from the generation branch (e.g., \texttt{blue} vs.\ \texttt{red}) are applied to the understanding branch, and we evaluate whether QA predictions can be shifted.
\textbf{(b) Layer-wise logit difference.} Here, $\Delta m_{\mathrm{LD}}$ quantifies the effect of steering vectors on model output logits. Understanding-derived vectors consistently shift model preference toward the target attribute, whereas generation-derived vectors show negligible or unstable effects.
\textbf{(c) Layer-wise subspace alignment analysis.} We measure alignment between steering vectors and the branch’s semantic subspace using the projection ratio (PSP), where higher values indicate better alignment. Generation-derived vectors exhibit low alignment with the understanding semantic subspace under $\text{gen} \rightarrow \text{und}$, indicating a semantic mismatch.
Here, $A \rightarrow B$ denotes learning on $A$ and applying to $B$ (self if $A=B$, cross otherwise).
}
\label{fig:ans}
\end{figure}

In Sec.~\ref{sec:und_to_gen}, we showed that steering vectors learned from the understanding branch can be transferred to the generation branch, enabling controllable image synthesis. In this section, we show that the reverse direction---from generation to understanding---does not work.

Similar to Sec.~\ref{sec:method}, we use semantic concepts constructed from our dataset \ourdata to study cross-branch transfer from generation branch into understand branch.
In this setting, as shown in Fig.~\ref{fig:ans}(a), contrastive pairs are formed using text prompts, \emph{e.g.}, $x_i^{+}:$ \texttt{``A blue car parked on a street.''} and $x_i^{-}:$ \texttt{``A red car parked on a street.''}. Using these contrastive pairs, we learn steering vectors from the generation branch based on VAE token representations and apply them to the understanding branch in a QA setting. More details are provided in Appendix~\ref{asec:gen_to_und}.

\paragraph{Generation Cannot Steer Understanding.} We first directly evaluate whether these vectors can change the model's textual predictions across multiple semantic concepts. As shown in Appendix~\ref{asec:fail_g2u}, the model's outputs remain unchanged under generation-derived steering, indicating that these vectors fail to induce meaningful changes in the model’s text outputs. To further examine this failure at a finer-grained logit level, we adopt the logit-difference metric $m_{\mathrm{LD}}$, which measures the model’s preference between a target answer and its contrastive alternative~\cite{logitdiff}. Specifically,
\begin{equation}
m_{\mathrm{LD}} = \mathrm{Logit}(\texttt{``positive''}) - \mathrm{Logit}(\texttt{``negative''}),
\end{equation}
where $\texttt{``positive''}$ and $\texttt{``negative''}$ tokens correspond to the target and contrastive concepts, respectively.
Then, we compare the change in $m_{\mathrm{LD}}$ with the steered model against the baseline (unsteered) model to quantify the effect of steering:
\begin{equation}
\Delta m_{\mathrm{LD}} = m_{\mathrm{LD}}^{\mathrm{steered}} - m_{\mathrm{LD}}^{\mathrm{baseline}}.
\end{equation}
A positive $\Delta m_{\mathrm{LD}}$ indicates that steering shifts the model toward the target concept, while values near zero suggest negligible behavioral change. A negative $\Delta m_{\mathrm{LD}}$ indicates that steering shifts the model toward the contrastive (opposite) concept.

In Fig.~\ref{fig:ans}(b), we report the layer-wise $\Delta m_{\mathrm{LD}}$ on the \texttt{`red'} vs.\ \texttt{`blue'} semantic concept, comparing steering vectors learned from the understanding branch and the generation branch under the same training data and steering scale.
We observe that steering vectors from understand branch consistently increase $\Delta m_{\mathrm{LD}}$, indicating effective control over the model’s output preference.
In contrast, vectors from generation branch yield near-zero or even negative $\Delta m_{\mathrm{LD}}$ across layers and exhibit unstable behavior, suggesting a failure to induce meaningful changes in the output logits.
This indicates that generation-derived steering cannot reliably shift the model toward the desired semantic direction in the understanding branch. To further verify that this phenomenon is not limited to a single semantic concept, we provide additional results across more semantic concepts in Appendix~\ref{asec:log_diff}.

\section{Why Is Transfer Asymmetric?}
\label{sec:why_asymmetry}

The results above reveal a clear asymmetry: steering succeeds from understanding to generation, but fails in the reverse direction. We now investigate \emph{why} this asymmetry arises through two complementary analyses.

\paragraph{Subspace Alignment Analysis.}
\label{sec:psp}
We hypothesize that semantic directions learned in the generation branch may not align with the semantic structure used by the understanding branch. To verify this, we introduce a PCA Subspace Projection (PSP) metric that measures how well a steering vector aligns with a branch’s dominant semantic directions (see Appendix~\ref{asec:psp} for details). Intuitively, a higher PSP indicates that the steering vector follows the model’s natural semantic structure, while a lower value suggests that it lies outside the space of meaningful semantics for that branch. We compute the PSP for the \texttt{red}–\texttt{blue} semantic direction in both self-branch and cross-branch settings. As shown in Fig.~\ref{fig:ans}(c), steering vectors from the understanding branch maintain high alignment when transferred to the generation branch, suggesting they capture generalized semantic directions. Conversely, generative steering vectors align poorly with the understanding branch, indicating their directions are not recognized as meaningful semantic variations. This discrepancy accounts for the observed asymmetry in cross-branch transferability.

\begin{figure}[!h]
\centering
\vspace{-0.3cm}
\includegraphics[width=0.98\linewidth]{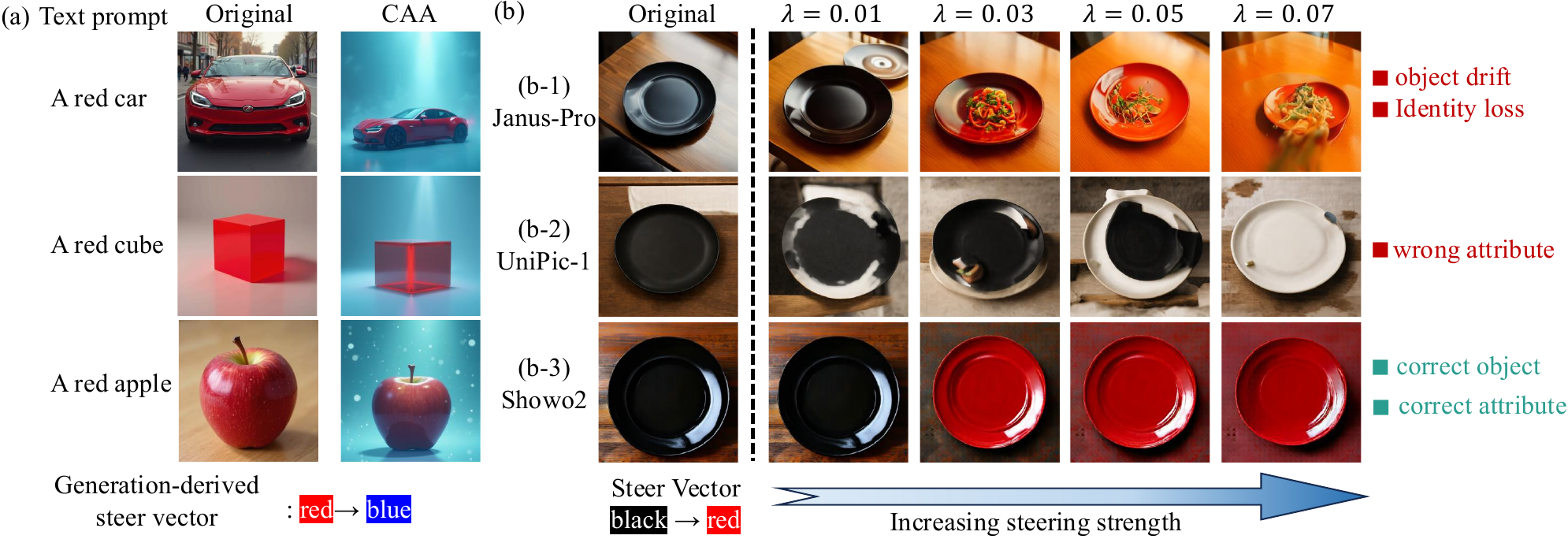}
\vspace{-4pt}
\caption{
\textbf{(a) Generation-derived vectors fail even in self-branch steering.} Steering vectors learned from the generation branch (red → blue) are applied back to the generation branch. Vectors instead alter global appearance (background, illumination) without changing the object.
\textbf{(b) Generalization across UMM architectures.}
Cross-branch steering fails in pure autoregressive structures (Janus-Pro~\cite{januspro}, UniPic-1~\cite{unipic}), but succeeds in hybrid AR+diffusion models (Show-o2~\cite{showo2}), highlighting the role of architecture.
}
\label{fig:gen2gen}
\end{figure}

\vspace{-0.2cm}
\paragraph{What do generation-derived vectors encode?}
To examine what information is encoded in steering vectors learned from the generation branch, we apply them back to the generation process in a self-branch setting. As shown in Fig.~\ref{fig:gen2gen}(a), the target semantic corresponds to object color (red $\rightarrow$ blue). While the intervention induces noticeable visual changes, it fails to change the object color to the target attribute. Instead, it primarily affects global appearance properties, such as background tone, illumination, and overall color distribution. This phenomenon suggests that generation-derived steering vectors mainly capture low-level, appearance-oriented features rather than fine-grained, object-level semantics. Consequently, they are unable to reliably steer the image toward the desired semantic target, which also explains their ineffectiveness in cross-branch transfer. This also confirms the PSP finding: generation vectors lie outside the understanding subspace precisely because they encode appearance-level rather than object-level semantics

\begin{tcolorbox}[colback=black!4, colframe=black!30, boxrule=0.6pt, arc=2pt, left=6pt, right=6pt, top=5pt, bottom=5pt, title={\textbf{Summary on asymmetric transfer}}]

The observed asymmetry may arise from differences in the information emphasized by the two branches. Understanding-derived vectors appear to capture more compositional, object-centric semantic directions that can transfer to the generation branch. In contrast, generation-derived vectors, under our current extraction protocol, seem to emphasize lower-level appearance statistics that are meaningful within the generation pipeline but are less aligned with the semantic structure used by the understanding branch. These results suggest that cross-branch transfer is more likely when the steering direction lies in a semantic subspace shared across branches.
\end{tcolorbox}

\section{Further Discussion}
\paragraph{Implication on UMM Architecture Design.}
\label{sec:umm_arch}
We further examine whether cross-branch steering generalizes across different UMM architectures. 
To this end, we evaluate two representative pure autoregressive (AR) models, Janus-Pro~\cite{januspro} and UniPic-1~\cite{unipic}, as well as a hybrid AR+diffusion model, Show-o2~\cite{showo2}, by transferring steering vectors from the understanding branch to the generation process (see Sec.~\ref{asec:umm_arch}). As shown in Fig.~\ref{fig:gen2gen}(b), cross-branch steering consistently fails on both AR models. For Janus-Pro~\cite{januspro}, increasing the steering strength leads to severe {object drift} and {identity loss}, where the generated object deviates from the original concept. 
For UniPic-1~\cite{unipic}, although the model performs strongly on standard generation and multimodal benchmarks, steering fails to induce the desired target semantic (\texttt{``red''} color), often resulting in incorrect or inconsistent attributes. In contrast, Show-o2~\cite{showo2}, a hybrid AR+diffusion model, successfully supports the transfer of steering vectors from the understanding branch to the generation process.

These results carry an important implication: cross-branch semantic unification is a property of architectural design, not a byproduct of strong individual capabilities. Hybrid AR+diffusion models, like BAGEL~\cite{bagel} and Show-o2~\cite{showo2}, operate in a shared continuous latent space with a shared backbone, naturally facilitating a unified semantic subspace;
pure AR models, which process visual and text tokens in separate codebook spaces, make it more difficult to achieve the same alignment, even when their benchmark performance is competitive or superior. This suggests that the community should look beyond task-level metrics when evaluating UMMs, and our cross-branch steering offers a complementary lens for assessing whether a model has achieved genuine representational unification.

\begin{figure}[!t]
\centering
\includegraphics[width=\linewidth]{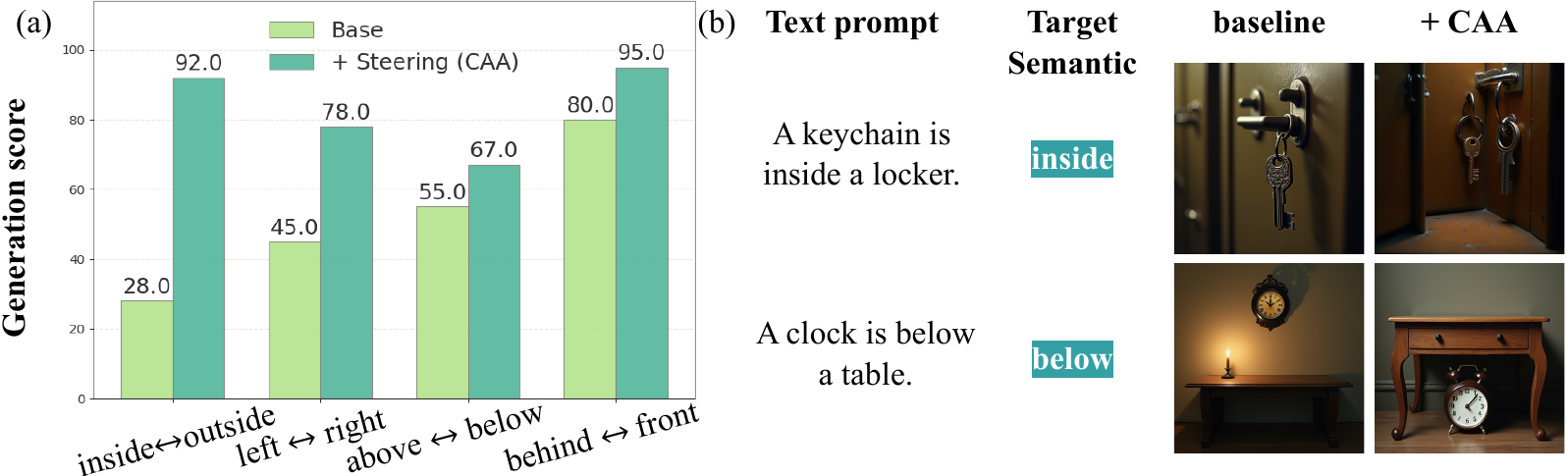}
\vspace{-3pt}
\caption{
\textbf{Understanding-derived steering reduces generation hallucination. } 
(a) Quantitative comparison shows that steering significantly improves generation on relational semantics. (b) Qualitative examples show corrected spatial configurations (e.g., inside, below) that the baseline model hallucinates.
}
\vspace{-4pt}
\label{fig:application}
\end{figure}

\paragraph{Application: Reducing Generation Hallucination in UMMs.} 
Here, we also explore whether understanding-derived steering vectors can mitigate image generation hallucinations, particularly on complex relational semantics where errors are most prevalent. We evaluate cross-branch steering on relational concepts (e.g., inside/outside, above/below) and report the \emph{generation score}, defined as the fraction of generated images that both contain the target object and satisfy the specified spatial relation. As shown in Fig.~\ref{fig:application}(a), CAA-based steering consistently improves generation scores across all relational concepts. Qualitative results in Fig.~\ref{fig:application}(b) further illustrate that steering enforces the desired relations, correcting common errors like incorrect spatial configurations.

\section{Conclusion}
In this work, we investigate whether multimodal understanding and generation in UMMs share a transferable semantic space via cross-branch semantic steering. We find that semantic directions learned from the understanding branch transfer effectively to generation, enabling controllable and faithful image synthesis. In contrast, generation-derived directions fail to influence understanding, revealing a fundamental asymmetry. Our analysis traces this asymmetry to a representational mismatch: understanding encodes object-centric semantics, while generation primarily captures low-level appearance features. These findings suggest that semantic alignment in UMMs depends critically on representation structure rather than model capability alone. Overall, cross-branch steering serves as both a diagnostic tool for probing latent semantics and a practical mechanism for improving generation faithfulness. Limitations are discussed in Appendix~\ref{asec:limitation}.
\newpage
\section*{Acknowledgments}
We thank Froilan Choi and Samuel Yeh for their valuable comments on the manuscript.
This work is supported in part by the AFOSR Young Investigator Program under award number FA9550-23-1-0184, National Science Foundation under awards IIS-2237037 and IIS-2331669, Schmidt Sciences Foundation, Open Philanthropy (now Coefficient Giving), Alfred P. Sloan Fellowship, UW-Madison Vilas Faculty Investigator Award, and gifts from Google and Amazon.

\bibliographystyle{unsrt}
\bibliography{neurips_2026.bib}

\newpage
\clearpage
\appendix
\startcontents[appendix]

\appendixtableofcontents

\clearpage

\section{Experiment Details}
\subsection{Implementation Detail}
\label{asec:implementation_details}
All experiments are conducted on NVIDIA A100 GPUs with 80GB memory, using two GPUs for most runs. We use officially released checkpoints for all models, including BAGEL~\cite{bagel}, Show-o2~\cite{showo2}, Janus-Pro~\cite{januspro},  and Qwen-VL-3 30B~\cite{qwen3vl}, without additional modification or re-training. Steering methods, including CAA~\cite{caa}, RepE~\cite{pca}, and ITI~\cite{cls}, are implemented in UMMs based on publicly available open-source implementations, following their standard formulations. 

For BAGEL generation, we use the default inference configuration unless otherwise specified.  Images are generated at a resolution of $1024$, with 50 generation steps and 50 diffusion timesteps. 
The text and image classifier-free guidance scales are both set to $1.0$. The CFG interval is set to $[0.4, 1.0]$, the CFG renormalization minimum is set to $0.0$, and the timestep shift is set to $3.0$. 
We use a random seed of 42 for all reported BAGEL experiments.

To ensure reproducibility, all experimental components rely on open-source methods and publicly accessible model checkpoints, and we provide detailed descriptions of dataset construction, steering procedures, and evaluation protocols in the main paper and appendix to facilitate independent verification.

\subsection{Dataset Detail}
\label{asec:data}
In this paper, we propose a contrastive paired dataset, \ourdata, for cross-branch steering.
Specifically, for each semantic type, we define multiple binary concepts, where each binary pair corresponds to a contrastive semantic direction, such as \texttt{red$\leftrightarrow$blue} and \texttt{above$\leftrightarrow$below}.
The complete set of contrastive semantic concepts is listed in Tab.~\ref{tab:data}.

\begin{figure}[!h]
\centering
\includegraphics[width=0.6\linewidth]{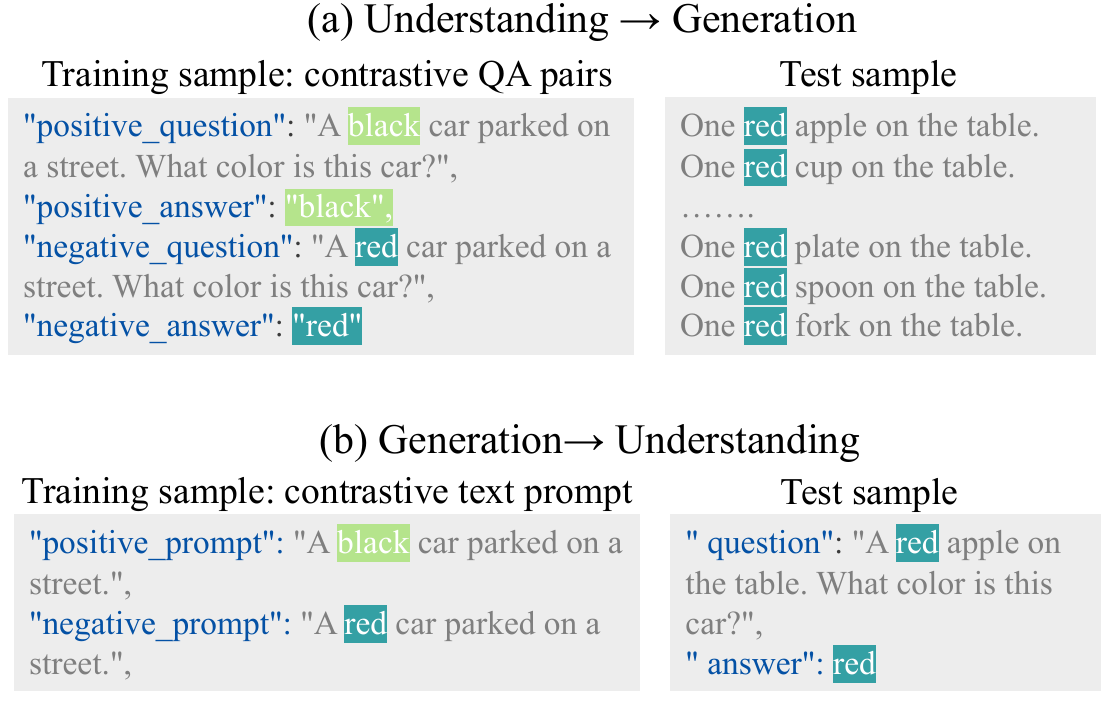}
\vspace{-6pt}
\caption{
\textbf{Illustration of cross-branch steering data samples.} 
(a) Understanding-to-generation (U→G): steering vectors are learned from contrastive QA pairs in the understanding branch and applied to image generation, where output image are evaluated using a VQA-based framework.
(b) Generation-to-understanding (G→U): steering vectors are learned from contrastive text prompt pairs in the generation branch and applied to the understanding task, where effectiveness is measured via logit-difference on target answers.
}
\label{fig:sample}
\end{figure}

For clarity, we also present examples of training and evaluation samples in Fig.~\ref{fig:sample}.
For training, each binary concept is expressed in two branch-specific forms.
For the understanding branch, we construct 50 contrastive question-answer pairs for each binary concept, where the two answers differ only in the target concept.
For the generation branch, we construct 50 paired text prompts, where the prompts differ only in the target semantic attribute.
These paired samples are used to extract branch-specific steering vectors.
For evaluation, we prepare 50 held-out samples for each individual semantic concept.
In U$\rightarrow$G steering, the held-out text prompts are used for image generation and then evaluated with a VQA-based framework.
In G$\rightarrow$U steering, the held-out QA samples are used to measure whether steering changes the model's answer preference toward the target concept.

\begin{table}[!h]
\centering
\small
\caption{\textbf{Hierarchical taxonomy of visually grounded contrastive concepts used for cross-branch steering.}
The concepts are organized into three levels: object-level (intrinsic properties), relational (spatial and structural relations), and scene-level (global appearance and style). Each concept is instantiated as a binary contrastive pair to enable precise semantic control.}
\label{tab:semantic_concepts}
\begin{tabular}{l l p{8cm}}
\toprule
\textbf{Category} & \textbf{Family} & \textbf{Example Binary / Contrastive Concepts} \\
\midrule

& \textbf{Color}
& black$\leftrightarrow$red, black$\leftrightarrow$white, blue$\leftrightarrow$green, pink$\leftrightarrow$red; 
  red$\leftrightarrow$blue, red$\leftrightarrow$green, yellow$\leftrightarrow$blue, yellow$\leftrightarrow$green \\
 \cmidrule(lr){2-3}
\multirow{2}{*}{\textbf{Object-level}}
& \textbf{Text} 
& apple$\leftrightarrow$orange, banana$\leftrightarrow$lemon, car$\leftrightarrow$bus, red$\leftrightarrow$blue, 
  red$\leftrightarrow$green, sherry$\leftrightarrow$monica, three$\leftrightarrow$ten, twenty-three$\leftrightarrow$ten \\
 \cmidrule(lr){2-3}
& \textbf{Counting} 
& one$\leftrightarrow$two, one$\leftrightarrow$three, one$\leftrightarrow$four, one$\leftrightarrow$five; 
  two$\leftrightarrow$three, two$\leftrightarrow$four, two$\leftrightarrow$five, 
  three$\leftrightarrow$four, three$\leftrightarrow$five, four$\leftrightarrow$five \\
\midrule
\multirow{2}{*}{\textbf{Relational"}} 
& \textbf{Position} 
& left$\leftrightarrow$right, above$\leftrightarrow$below, above$\leftrightarrow$under, inside$\leftrightarrow$outside, 
  front$\leftrightarrow$behind, near$\leftrightarrow$far, clustered$\leftrightarrow$scattered \\
\midrule
& \textbf{Appearance} 
& clean$\leftrightarrow$dirty, full$\leftrightarrow$empty, intact$\leftrightarrow$broken, new$\leftrightarrow$worn, 
  round$\leftrightarrow$square, smooth$\leftrightarrow$rough, tall$\leftrightarrow$short, wooden$\leftrightarrow$metal \\
 \cmidrule(lr){2-3}
\multirow{3}{*}{\textbf{Scene-level}}  & \textbf{Style \& Tone} 
& colorful$\leftrightarrow$monochrome, vivid$\leftrightarrow$dull, warm$\leftrightarrow$cool, 
  realistic$\leftrightarrow$cartoon, sketch$\leftrightarrow$oil painting, 
  happy$\leftrightarrow$unhappy, peaceful$\leftrightarrow$chaotic, romantic$\leftrightarrow$gloomy \\
 \cmidrule(lr){2-3}
& \textbf{Persona} 
& extraversion$\leftrightarrow$introversion, neuroticism$\leftrightarrow$ calmness\\
\bottomrule
\label{tab:data}
\end{tabular}
\end{table}

\subsection{From Generation to Understanding Steering Process Detail}
\label{asec:gen_to_und}

\begin{figure}
    \centering
    \includegraphics[width=0.8\linewidth]{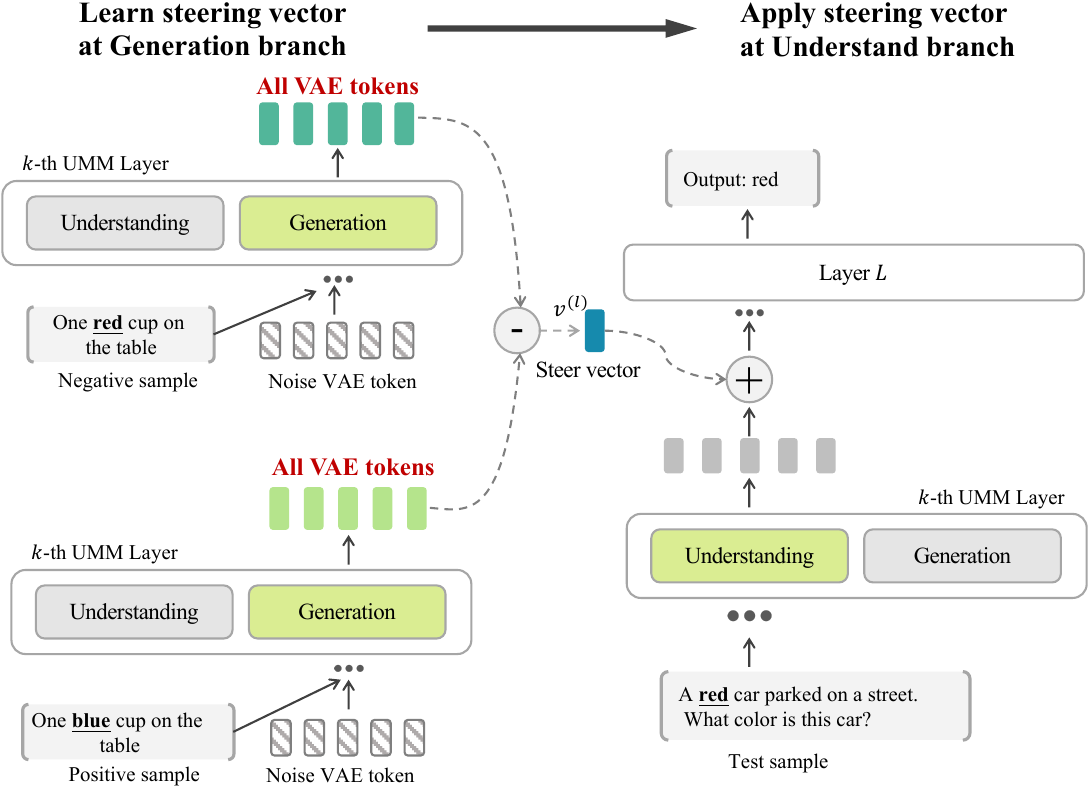}
    \vspace{-5pt}
\caption{
\textbf{Cross-branch steering pipeline (Gen $\rightarrow$ Und).} First, contrastive text prompt pairs (e.g., \texttt{`blue'} vs.\ \texttt{`red'}) are fed through the generation branch. At each layer, we extract residual-stream activations from the VAE-token hidden states and use them to compute a generation-derived steering vector. At inference time, a test sample is fed into the understanding branch, and the steering vector is added to the text-token hidden states at each layer to test whether generation-derived directions can bias the model's textual prediction.
} 
\label{fig:gen2und}
\end{figure}

Fig.~\ref{fig:gen2und} illustrates the pipeline for transferring steering vectors from the generation branch to the understanding branch. This setting is the reverse direction of the Und$\rightarrow$Gen transfer studied in the main text. The goal is to examine whether semantic directions learned from the generation process can also influence the model's textual understanding behavior.

Given a pair of contrastive generation prompts that differ only in the target semantic concept, such as \texttt{`red'} versus \texttt{`blue'}, we feed each prompt into the generation branch and collect the hidden states of the VAE latent tokens at each UMM layer. Unlike the Und$\rightarrow$Gen setting, where the steering vector is learned from the text hidden state at the answer-token position in the understanding branch, the generation branch does not have a single answer token that directly represents the target semantic label. Therefore, we use global mean pooling over VAE latent tokens as an architecture-agnostic extraction strategy. This choice avoids introducing additional external grounding modules, such as object detectors, segmentation masks, or attention-based token selection heuristics, which may themselves introduce model- or category-specific biases (More details refer to ~\ref{asec:object_centric_extraction}).
Specifically, for each contrastive prompt pair $x_i=(x_{+,i},x_{-,i})$ at layer $l$, we compute the mean residual-stream activation over all VAE latent token positions:
\begin{equation}
    \bar{\mathbf{h}}^{(l)}_{\mathrm{gen,+}}(x_i)
    =
    \frac{1}{|\mathcal{T}_{\mathrm{vae}}|}
    \sum_{t \in \mathcal{T}_{\mathrm{vae}}}
    \mathbf{h}^{(l)}_{t,+}, \quad
    \bar{\mathbf{h}}^{(l)}_{\mathrm{gen,-}}(x_i)
    =
    \frac{1}{|\mathcal{T}_{\mathrm{vae}}|}
    \sum_{t \in \mathcal{T}_{\mathrm{vae}}}
    \mathbf{h}^{(l)}_{t,-},
\end{equation}
where $\mathcal{T}_{\mathrm{vae}}$ denotes the set of VAE latent token positions and $\mathbf{h}^{(l)}_{t,\pm}$ denotes the residual-stream hidden state of token $t$ at layer $l$ under the positive or negative generation prompt. For notational simplicity, we denote these averaged activations as
$\mathbf{h}^{(l)}_{+,i}=\bar{\mathbf{h}}^{(l)}_{\mathrm{gen,+}}(x_{+,i})$ and
$\mathbf{h}^{(l)}_{-,i}=\bar{\mathbf{h}}^{(l)}_{\mathrm{gen,-}}(x_{-,i})$.
This yields a set of paired generation-branch activations
$\{(\mathbf{h}^{(l)}_{+,i}, \mathbf{h}^{(l)}_{-,i})\}_{i=1}^N$,
which are then used to compute layer-wise steering vectors with the methods described in Sec.~\ref{asec:steeringllm}.

At inference time, we transfer the generation-derived vector to the understanding branch. Given an input question-answering sample, the model processes the text input through the understanding branch. At each layer, we add the generation-derived steering vector $\mathbf{v}^{(l)}$ to the hidden states of the text tokens:
\begin{equation}
    \mathbf{h}^{(l)}_{t}
    \leftarrow
    \mathbf{h}^{(l)}_{t}
    +
    \alpha \mathbf{v}^{(l)},
    \qquad
    t \in \mathcal{T}_{\mathrm{text}},
\end{equation}
where $\alpha$ is the steering strength and $\mathcal{T}_{\mathrm{text}}$ denotes the text-token positions in the understanding input. We then evaluate whether the intervention shifts the model's answer toward the target semantic concept.

\subsection{LLM Steering Methods Detail}
\label{asec:steeringllm}
In Sec.~\ref{sec:steer_learn}, we briefly introduced three representative methods for estimating steering directions: CAA~\cite{caa}, RepE~\cite{pca}, ITI~\cite{cls}. Here, we provide their formal definitions in details.

For a target concept $k$, we construct contrastive pairs from $\mathcal{D}^k$ and feed them into the source branch. We extract residual stream activations at layer $l$, taken from the answer token position in the understanding branch, or from VAE token positions in the generation branch.
This yields a set of paired activations
$\{\mathbf{h}^{(l)}_{+,i}, \mathbf{h}^{(l)}_{-,i}\}_{i=1}^N$,
where $+$ and $-$ denote the positive and negative instances, respectively.
We define the difference vectors as $\Delta \mathbf{h}_i^{(l)} = \mathbf{h}^{(l)}_{+,i} - \mathbf{h}^{(l)}_{-,i}$.

\paragraph{(1) Mean Difference (CAA~\cite{caa}).}
The steering direction is estimated as the average contrastive shift:
\begin{equation}
\mathbf{v}^{(l)}_{\text{CAA}} = \frac{1}{N} \sum_{i=1}^{N} \Delta \mathbf{h}_i^{(l)}.
\end{equation}

\paragraph{(2) First Principal Component of the Difference Vector (RepE~\cite{pca}).}
We compute the top principal component of the difference vectors:
\begin{equation}
\mathbf{v}^{(l)}_{\text{RepE}} = \mathrm{TopPC}\left(\{\Delta \mathbf{h}_i^{(l)}\}_{i=1}^N\right),
\end{equation}
where $\mathrm{TopPC}(\cdot)$ returns the leading principal component (unit-normalized) corresponding to the largest singular value.

\paragraph{(3) Normal Vector of a Learned Decision Boundary (ITI~\cite{cls}).}
We train a linear classifier to distinguish positive and negative activations, and use the normal vector of the decision boundary as the steering direction:
\begin{equation}
\mathbf{v}^{(l)}_{\text{ITI}} = \mathrm{Classify}\left(\{ \mathbf{h}^{(l)}_{\pm,i} \}_{i=1}^N \right),
\end{equation}
where $\mathrm{Classify}(\cdot)$ returns the learned separating direction, normalized to match the standard deviation along that axis.

In summary, CAA~\cite{caa} captures the average semantic shift between contrastive states, RepE~\cite{pca} extracts the dominant variation direction, and ITI~\cite{cls} explicitly optimizes discriminative separation.
Despite their different formulations, all three methods recover a semantic direction $\mathbf{v}^{(l)}$ in hidden space, which can be used to steer model behavior toward the target concept.

\subsection{Evaluation Metric}
\label{asec:eval_metric}

\begin{figure}[t]
\centering
\begin{tcolorbox}[
    enhanced,
    width=0.9\linewidth,
    colback=white,
    colframe=black,
    boxrule=0.8pt,
    arc=2pt,
    left=10pt,
    right=10pt,
    top=10pt,
    bottom=10pt,
    title=\textbf{Steering Effectiveness Evaluation Prompt Template},
    coltitle=black,
    colbacktitle=white,
    fonttitle=\bfseries,
    attach boxed title to top left={
        xshift=2pt,
        yshift=-3mm
    },
    boxed title style={
        colback=white,
        colframe=black,
        boxrule=0.8pt,
        arc=2pt,
        left=5pt,
        right=5pt,
        top=2pt,
        bottom=2pt
    }
]
\small
You are judging whether an image satisfies a target semantic induced by steering.
Use only visible evidence from the image. Return JSON only and do not reveal chain-of-thought.
\vspace{0.6em}
\noindent\textbf{Target semantic:} \\
\textit{[Target Semantic $s$, e.g., blue apple, two dogs, text ``STOP'', object above table]}

\vspace{0.6em}
\noindent\textbf{Judgment mode:} \\
\textit{[Judge Mode $m \in \{$item\_attribute, count, text\_content, persona, relation, global\_attribute$\}$]}

\vspace{0.6em}
\noindent\textbf{Stage 1: Evidence / object-presence check} \\
Question ID: \texttt{q\_gate} \\
Question Type: \textit{[Gate Type]} \\
Question: \textit{[Does the image clearly contain the target item / readable text / person / required relation objects / visual evidence?]} \\
Intent: Check whether the image contains sufficient visible evidence for evaluating the target semantic. \\
Positive signal: \textit{[The required object, subject, text, or visual evidence is clearly visible.]}

\vspace{0.6em}
\noindent\textbf{Stage 2: Target-semantic match check} \\
Question ID: \texttt{q\_target} \\
Question Type: \textit{[Target Type]} \\
Question: \textit{[If the required evidence is present, does it match the target semantic $s$?]} \\
Intent: Check whether the generated image satisfies the desired steered semantic. \\
Positive signal: \textit{[The visible evidence matches the target attribute/count/text/persona/relation/global semantic.]}

\vspace{0.6em}
\noindent\textbf{Return JSON only:}
\begin{verbatim}
{
  "question_id": "...",
  "question_type": "...",
  "label": "yes" | "no" | "uncertain",
  "confidence": 0.0,
  "answer": "...",
  "evidence": "...",
  "short_rationale": "..."
}
\end{verbatim}
\end{tcolorbox}
\vspace{-10pt}
\caption{
Two-stage VQA evaluation prompt for measuring steering success.
The judge first verifies whether the generated image contains sufficient visible evidence for evaluation,
and then checks whether the target semantic is satisfied.
A generation is counted as successful only when both the gate question and the target-semantic question are answered \texttt{yes}.
}
\label{fig:vqa_eval_prompt}
\end{figure}
\paragraph{Steering Success Rate (SSR).}
Following Sec.~\ref{sec:method}, we adopt a question-answering (QVA) protocol with Qwen-VL-3 30B~\cite{qwen3vl} as the evaluator to assess steering effectiveness in the generation branch.
The evaluation consists of two criteria. 
First, we verify whether the generated image preserves the target object, ensuring that steering does not alter object identity.
Second, we assess whether the desired semantic attribute (e.g., spatial relation or color) is successfully enforced.
A sample is considered successful only if both criteria are satisfied. We show the two-stage VQA evaluation prompt template in Fig.~\ref{fig:vqa_eval_prompt}. Formally, we define the \emph{steering success rate} (SSR) as:
\begin{equation}
\text{SSR} = \frac{1}{N} \sum_{i=1}^{N} \mathbb{I}\big( y_i^{\text{obj}} = 1 \;\land\; y_i^{\text{attr}} = 1 \big),
\end{equation}
where \(N\) is the total number of evaluation samples,
\(y_i^{\text{obj}} \in \{0,1\}\) indicates whether the target object is preserved,
and \(y_i^{\text{attr}} \in \{0,1\}\) indicates whether the desired semantic attribute is successfully enforced.
\(\mathbb{I}(\cdot)\) denotes the indicator function. Here, we use Qwen-VL-3 30B~\cite{qwen3vl} as the evaluator due to its strong performance on multimodal reasoning benchmarks. In addition, our evaluation is based on simple, visually grounded verification (e.g., object presence and attribute matching), which reduces reliance on complex reasoning and improves robustness. 

\begin{figure}[!t]
\centering
\includegraphics[width=\linewidth]{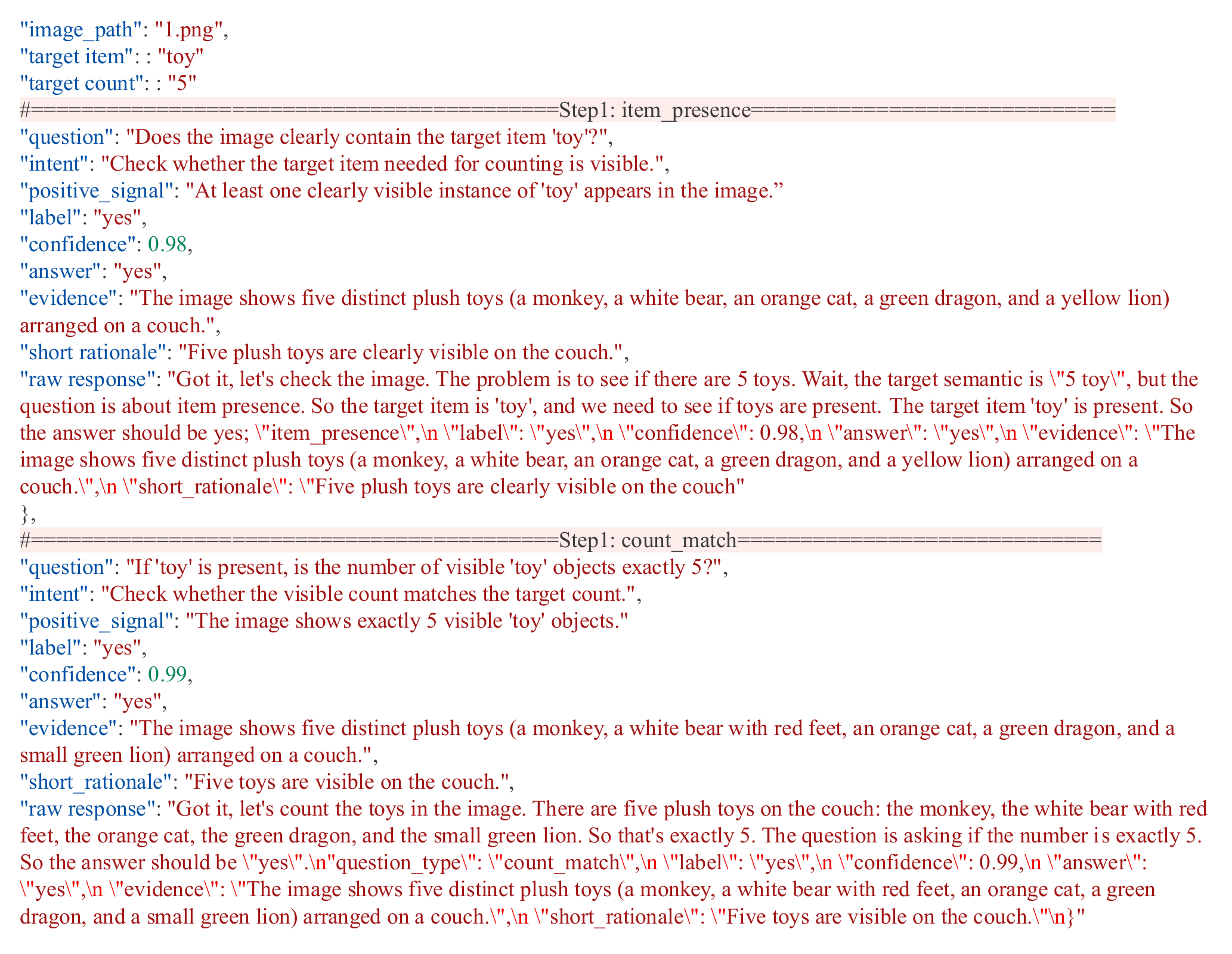}
\caption{
\textbf{Qualitative examples of evaluator outputs from Qwen3 VL-30B~\cite{qwen3vl}.} 
}
\label{fig:eval_sample}
\end{figure}

Here, we include qualitative examples of evaluator outputs in Fig.~\ref{fig:eval_sample} for clarity. 

\paragraph{Generation Score.} Following GenEval~\cite{geneval} and WISE~\cite{niu2025wise}, we adopt a VQA-based evaluation protocol using Qwen3-VL-30B~\cite{qwen3vl} as the judge to assess text--image alignment.
A generated image is considered correct only if it contains the target object and satisfies the target attribute or relation specified in the prompt. Formally, we define the \emph{generation score} as:
\begin{equation}
\text{GenScore} = \frac{1}{N} \sum_{i=1}^{N} \mathbb{I}\big( y_i^{\text{obj}} = 1 \;\land\; y_i^{\text{attr}} = 1 \big),
\end{equation}
where \(N\) is the total number of evaluation samples,
\(y_i^{\text{obj}} \in \{0,1\}\) indicates whether the target object is present in the generated image,
and \(y_i^{\text{attr}} \in \{0,1\}\) indicates whether the desired attribute or relation is satisfied.
\(\mathbb{I}(\cdot)\) denotes the indicator function. The evaluation prompt for Qwen3-VL-30B~\cite{qwen3vl} follows the same two-stage structure as in SSR, consisting of object presence verification and attribute (or relation) matching. The key difference is that, in SSR, the target attribute is defined by the applied steering vector, whereas in the generation score it is specified by the input prompt.

\section{Additional results}
\subsection{Human Agreement Study}
\label{asec:human_agreement}

To validate the reliability of our question-answering-based visual assessment (QVA) protocol, we conduct a human agreement study on a randomly sampled subset of generated images.
Specifically, we sample 50 images for each of the seven semantic types in UMMSteer: \textit{color}, \textit{text}, \textit{counting}, \textit{position}, \textit{style}, \textit{appearance}, and \textit{persona}, resulting in 350 annotated images in total.
For each image, a human annotator independently judges the same binary verification criteria used by the automatic evaluator: whether the target object is present and whether the target semantic attribute or relation is correctly realized.
The annotator is shown the generated image, the prompt, the target object, and the target semantic attribute or relation.

We compute Cohen's $\kappa$ between the human annotations and Qwen-VL-3 30B~\cite{qwen3vl} predictions to quantify chance-corrected human--model agreement.
We also report raw agreement for completeness.
For each sample, we evaluate overall correctness, where a generated image is considered correct only if both the target object is present and the target attribute or relation is correct.
Cohen's $\kappa$ accounts for chance agreement and is therefore more conservative than raw agreement.

\begin{table}[t]
\centering
\caption{
Human--model agreement for validating the QVA-based evaluator.
We report raw agreement and Cohen's $\kappa$ between human annotations and Qwen-VL-3 30B predictions across semantic types.
}
\label{tab:human_model_agreement}
\begin{tabular}{lcc}
\toprule
Semantic type & Agreement (\%) & Cohen's $\kappa$ \\
\midrule
Text & 98.0 & 0.658 \\
Position & 98.0 & 0.790 \\
Style & 100.0 & 1.000 \\
Persona & 90.0 & 0.800 \\
Appearance & 90.0 & 0.786 \\
Counting & 98.0 & 0.929 \\
Color & 94.0 & 0.540 \\
\midrule
Overall & 95.4 & 0.894 \\
\bottomrule
\end{tabular}
\end{table}

As shown in Tab.~\ref{tab:human_model_agreement}, Qwen-VL-3 30B achieves high agreement with human annotations across all seven semantic types.
Across 350 annotated samples, the QVA-based evaluator obtains a raw agreement of 95.4\% and a Cohen's $\kappa$ of 0.894, indicating strong chance-corrected agreement with human judgments.
These results support the reliability of our automatic evaluation protocol.

\subsection{Full Steer Effectiveness Evaluation for Understanding-to-Generation Steering}
\label{asec:steer_effective}
\begin{figure}[!t]
\centering
\includegraphics[width=0.85\linewidth]{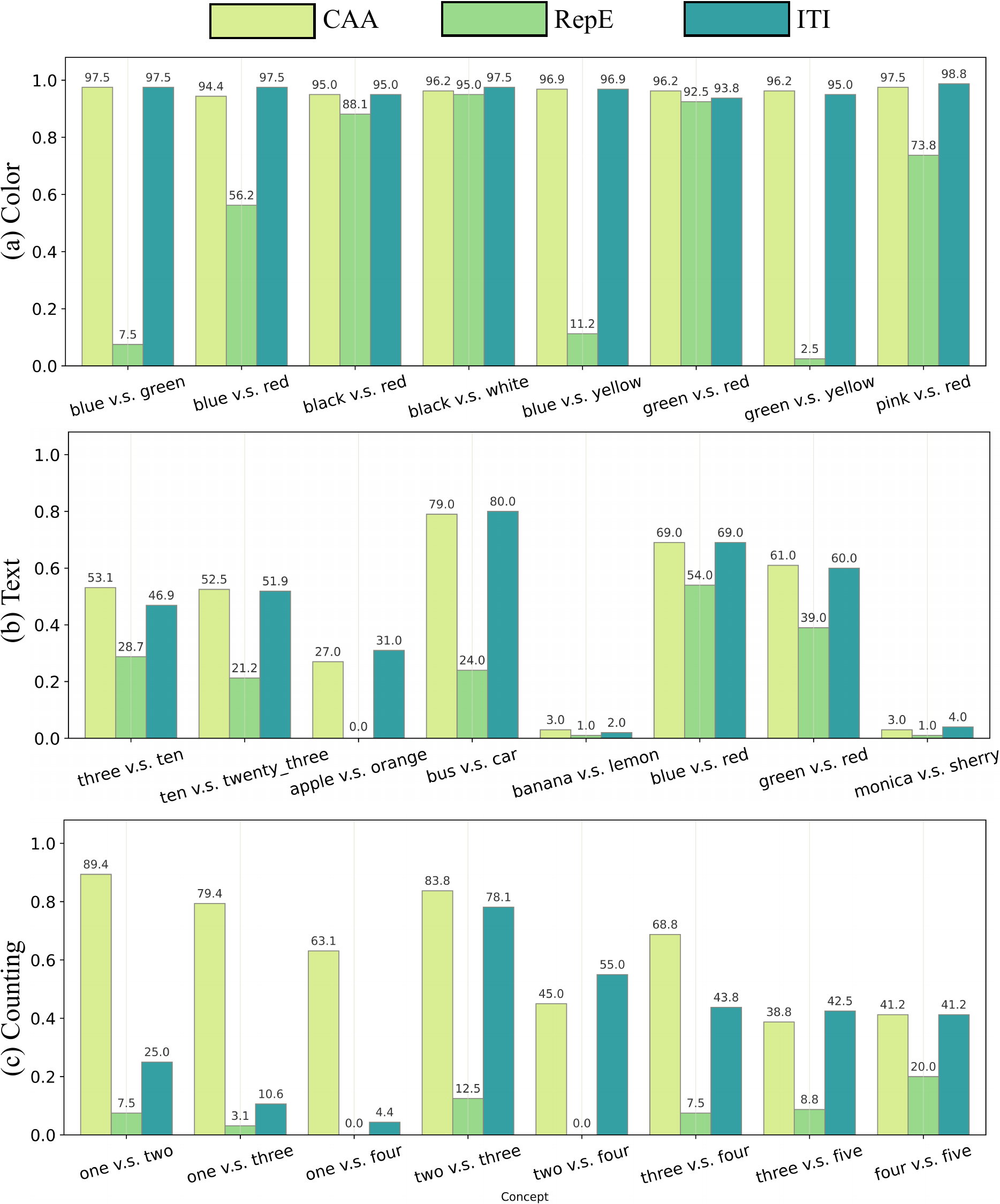}
\caption{Steering success rates for each concept across object-level semantics.}
\label{fig:object}
\end{figure}

\begin{figure}[!t]
\centering
\includegraphics[width=0.8\linewidth]{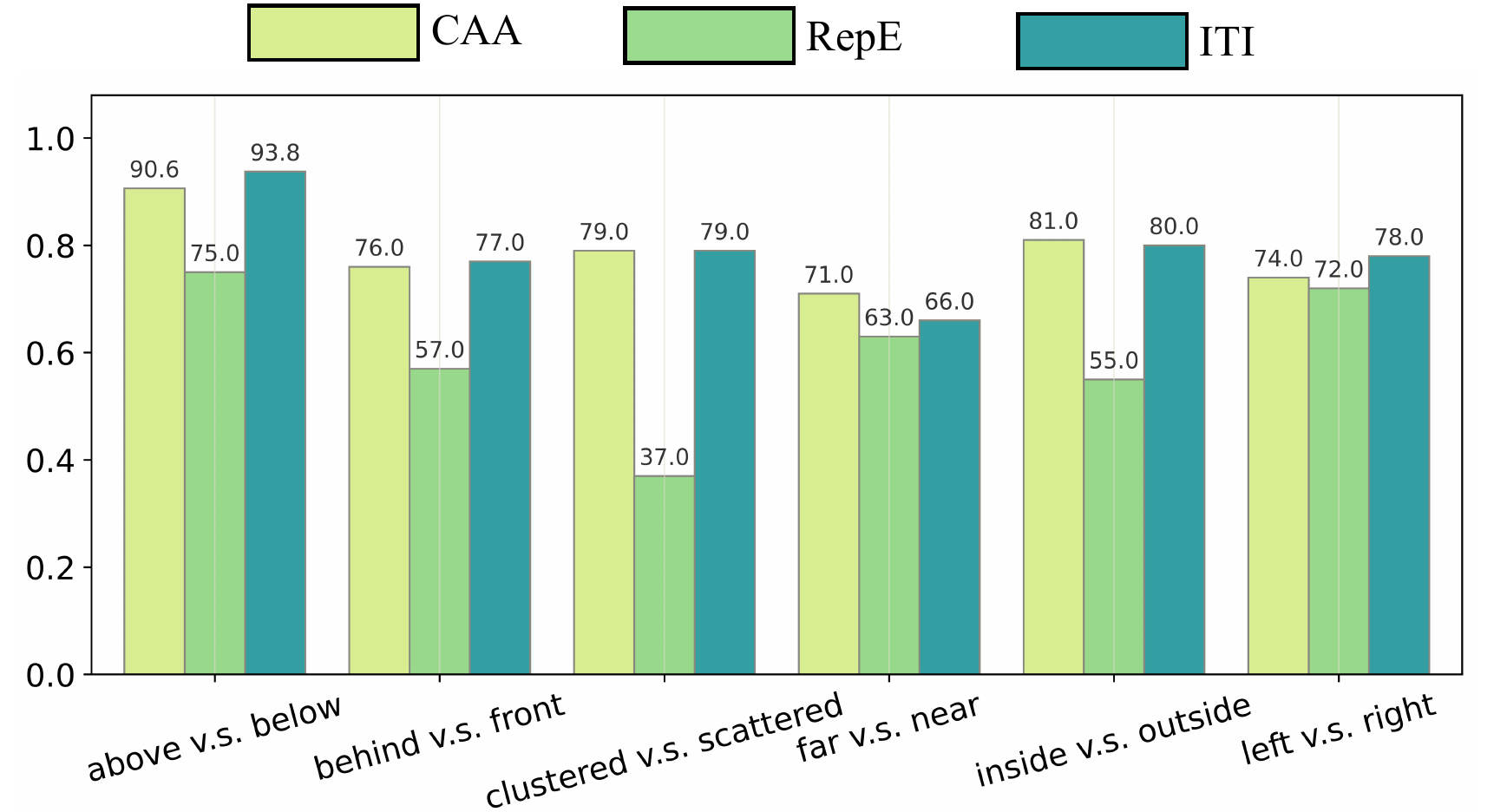}
\caption{Steering success rates for each concept across relational semantics.
}
\label{fig:rational}
\end{figure}

\begin{figure}[!t]
\centering
\includegraphics[width=\linewidth]{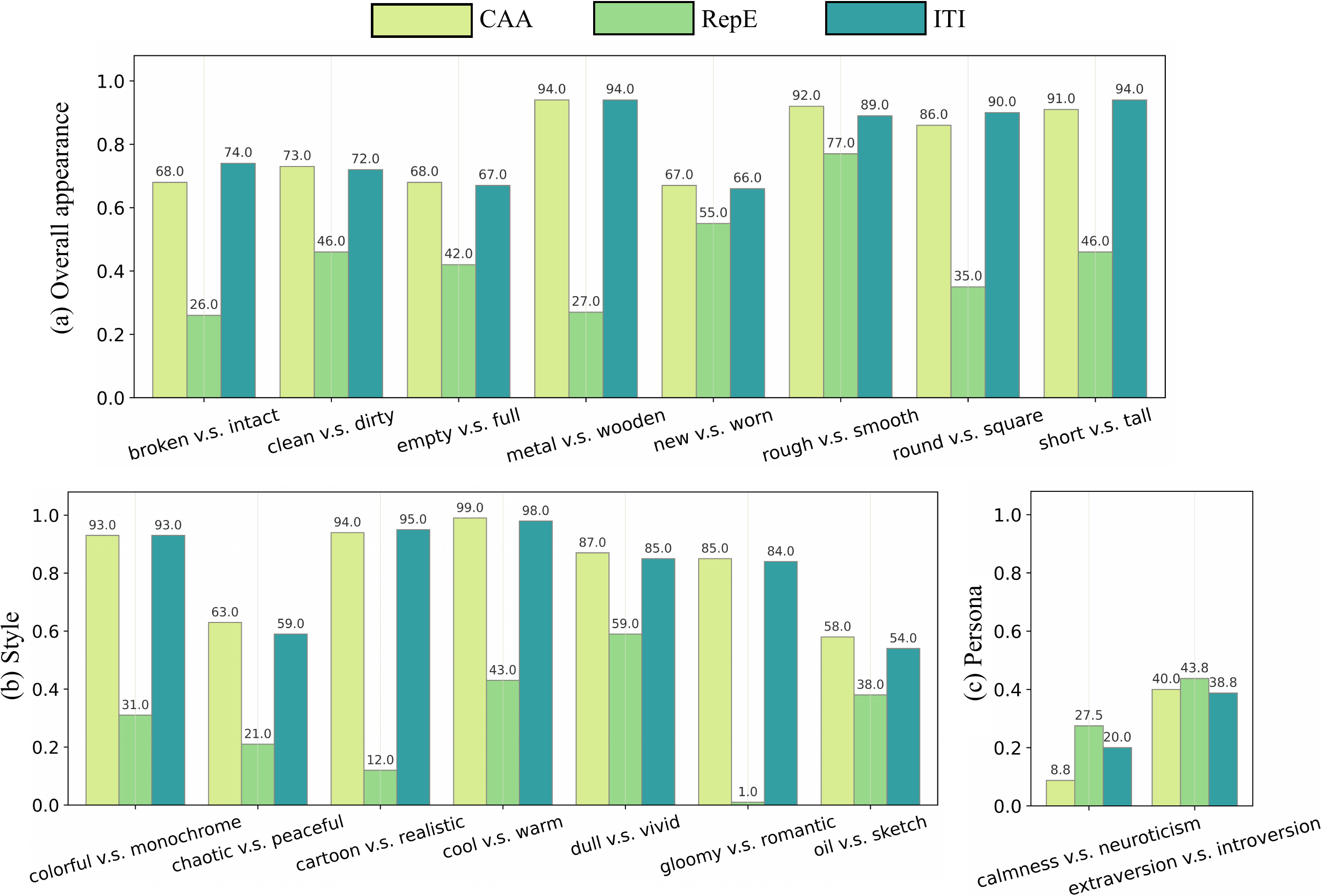}
\vspace{-5pt}
\caption{Steering success rates for each concept across scene-level semantics.
}
\label{fig:sence}
\end{figure}

We observe substantial within-category variance in Tab.~\ref{tab:u2g}, and therefore further present the full steering success rate (SSR) of all semantic concept pair from our \ourdata in Fig.~\ref{fig:object}, Fig.~\ref{fig:sence} and Fig.~\ref{fig:rational}. Here, each concept pair (\texttt{`a'} vs. \texttt{`b'}) is evaluated bidirectionally: (i) for \texttt{`a'}$\rightarrow$\texttt{`b'}, we use prompts containing attribute \texttt{`a'} and apply the \texttt{`a'}$\rightarrow$\texttt{`b'} steering vector to enforce attribute \texttt{`b'}; (ii) for \texttt{`b'}$\rightarrow$\texttt{`a'}, we perform the reverse. The final SSR is obtained by averaging the two directions.
 
First, we report the SSR for all \textbf{object-level} semantic concepts in Fig.~\ref{fig:object}. For \textit{Color} semantics, both CAA~\cite{caa} and ITI~\cite{cls} achieve consistently high performance across all concept pairs, indicating that these methods can reliably capture and transfer color-related semantic directions. In contrast, RepE~\cite{pca} exhibits unstable behavior, particularly failing on concepts involving \texttt{`green'}, \texttt{`yellow'}, and \texttt{`blue'}, leading to significantly degraded performance. For {\textit{Text}} semantics, all methods perform well on short and simple tokens (e.g., \texttt{`apple'}, \texttt{`orange'}), suggesting that such semantics are relatively easy to encode and transfer. However, performance drops substantially for longer or more complex words (e.g., \texttt{`banana'}, \texttt{`lemon'}, \texttt{`Monica'}, \texttt{`Sherry'}), where all methods struggle, indicating limitations in capturing fine-grained textual semantics in generation. For {\textit{Counting}} semantics, we observe a clear trend that steering difficulty increases with the target number. Specifically, concepts involving larger counts (e.g., 4 vs. 5) yield much lower SSR compared to smaller counts (e.g., 1 vs. 2), suggesting that precise numerical control in generation remains challenging, especially as the counting complexity increases.

Second, we report the full SSR for \textbf{relational} semantics in Fig.~\ref{fig:rational}.
Most spatial relations (e.g., \texttt{`above'} vs. \texttt{`below'}, \texttt{`inside'} vs. \texttt{`outside'}, \texttt{`left'} vs. \texttt{`right'}) can be reliably steered, with CAA~\cite{caa} and ITI~\cite{cls} achieving consistently strong performance. However, more complex structural relations such as \texttt{`clustered'} vs. \texttt{`scattered'} are noticeably harder, especially for RepE~\cite{pca}, which shows large performance drops. Overall, relational semantics are easier to steer than counting and complex text, but still exhibit variability depending on spatial complexity.

Last, we report the full SSR for \textbf{scene-level} semantics in Fig.~\ref{fig:sence}.
For \textit{Appearance}, CAA~\cite{caa} and ITI~\cite{cls} achieve strong performance on most concepts (e.g., \texttt{`metal'}, \texttt{`wooden'}, \texttt{`round'}, \texttt{`square'}, \texttt{`short'}, \texttt{`tall'}), while some attributes such as \texttt{`clean'}, \texttt{`dirty'}, \texttt{`empty'} and \texttt{`full'} are relatively harder. RepE~\cite{pca} shows unstable behavior and significantly underperforms across many pairs.
For \textit{Style}, CAA~\cite{caa} and ITI~\cite{cls} perform well on visually distinctive styles such as \texttt{`cartoon'}, \texttt{`realistic'}, \texttt{`colorful'}, \texttt{`monochrome'}, and \texttt{`cool'}, \texttt{`warm'}. The lower SSR on \texttt{`oil'}, \texttt{`sketch'} is likely due to the difficulty of generating the oil-painting style itself.
For \textit{Persona}, all methods perform poorly (e.g., \texttt{`calmness'}, \texttt{`neuroticism'}), suggesting that such abstract and weakly visualizable semantics are difficult to steer.

\subsection{Full Steer Effectiveness Evaluation for Generation-to-Understanding Steering}
\label{asec:fail_g2u}
We further evaluate whether steering vectors learned from the generation branch can be transferred to the understanding branch.

Specifically, for each contrastive semantic concept, such as \texttt{`blue'} vs.\ \texttt{`red'}, we extract steering vectors from the generation branch and apply them to the hidden states of the understanding branch during visual question answering. The goal is to test whether these generation-derived directions can shift the model's textual prediction toward the target semantic label, as shown in Fig.~\ref{fig:ans}(a).

For evaluation, we use the target semantic label as the ground-truth answer and compute the prediction accuracy between the model output and the target semantic label. For example, when applying a \texttt{red}$\rightarrow$\texttt{blue} steering direction, a successful transfer should make the model answer \texttt{blue} when queried about the corresponding visual attribute with an input \texttt{`a red car'}. 

To rule out the possibility that the failure is caused by an inappropriate steering magnitude, we conduct a dense sweep over the steering scale $\alpha$ and evaluate the model under each scale value. However, even under this exhaustive scale search, we do not observe any scale value that consistently shifts the model output toward the target semantic label. Across all evaluated semantic concepts, \textbf{the prediction accuracy remains zero}, indicating that generation-derived steering vectors fail to induce meaningful changes in the understanding branch. This result shows that semantic directions learned from the generation branch are not directly usable for controlling text-side understanding behavior.

\subsection{Full Evaluation of Understanding-to-Generation Steering across UMM Architectures}
\label{asec:umm_arch}

In Sec.~\ref{sec:umm_arch}, we examine whether cross-branch steering generalizes across different unified multimodal model (UMM) architectures. Here, we provide additional quantitative results for understanding-to-generation (Und$\rightarrow$Gen) steering on Show-o2~\cite{showo2}, Janus-Pro~\cite{januspro}, and UniPic-1~\cite{unipic}. We use CAA~\cite{caa} to extract the steering vectors and follow the same steering success rate (SSR) evaluation protocol used for BAGEL~\cite{bagel}. The BAGEL results are included for reference.

\begin{table}[t]
    \centering
    \small
    \caption{
    Und$\rightarrow$Gen steering success rate (SSR, \%) across UMMs with different generation architectures. All steering vectors are extracted using CAA.
    }
    \label{tab:umm_arch_und_to_gen}
    \begin{tabular}{llcccc}
        \toprule
        \textbf{Model} &
        \textbf{Architecture} &
        \textbf{Color} &
        \textbf{Counting} &
        \textbf{Position} &
        \textbf{Style} \\
        \midrule
        BAGEL     & Hybrid AR+diffusion & \textbf{96.20} & \textbf{68.75} & \textbf{78.60} & \textbf{82.71} \\
        Show-o2   & Hybrid AR+diffusion & 87.13 & 43.07 & 32.20 & 64.20 \\
        Janus-Pro & Pure AR             & 48.75 & 20.57 & 8.67  & 47.80 \\
        UniPic-1  & Pure AR             & 2.30  & 13.07 & 13.40 & 37.80 \\
        \bottomrule
    \end{tabular}
\end{table}

As shown in Table~\ref{tab:umm_arch_und_to_gen}, BAGEL achieves the highest SSR across all four semantic categories. Show-o2 also exhibits substantial Und$\rightarrow$Gen steering effects, although its performance is lower than that of BAGEL, particularly for counting and position. In contrast, the two evaluated pure autoregressive models, Janus-Pro and UniPic-1, generally achieve substantially lower SSR, consistent with the qualitative results reported in Fig.~\ref{fig:gen2gen}(b).

These results provide additional evidence that, under our current steering setup, Und$\rightarrow$Gen transfer is more effective in the evaluated hybrid AR+diffusion models than in the evaluated pure autoregressive models. A possible explanation is that our intervention directly adds steering vectors to continuous VAE hidden states during generation, making the current protocol naturally more compatible with models such as BAGEL and Show-o2. Therefore, these results should be interpreted as evidence of compatibility between the proposed steering protocol and continuous VAE representations, rather than as evidence that hybrid architectures inherently possess more unified semantic representations.

\section{Additional Ablation Study and Analysis}

\subsection{Additional Baseline Experiments}
\label{asec:additional_baselines}

To further examine whether successful cross-branch steering can be attributed to a generic perturbation rather than the extracted target direction, we introduce two additional baselines for understanding-to-generation (Und$\rightarrow$Gen) steering on BAGEL~\cite{bagel}:
\begin{itemize}
    \item \textit{Random-direction baseline.}
At each layer, we sample a random direction and rescale it to match the $\ell_2$ norm of the corresponding understanding-derived CAA~\cite{caa} vector. This baseline tests whether a perturbation with a comparable magnitude is sufficient to induce successful steering.
   \item \textit{Unrelated-direction baseline.}
We extract an understanding-derived CAA vector from a semantically unrelated persona concept, specifically \texttt{extraversion} versus \texttt{introversion}, and apply it to the evaluation of color, counting, position, and style. This baseline tests whether an unrelated but semantically meaningful direction can produce similar steering effects.
\end{itemize}

All directions are applied using the same intervention and evaluation protocol. We report the steering success rate (SSR) in Table~\ref{tab:additional_baselines}.

\begin{table}[t]
    \centering
    \small
    \caption{
    Und$\rightarrow$Gen steering success rate (SSR, \%) on BAGEL using CAA and two additional baselines.
    }
    \label{tab:additional_baselines}
    \begin{tabular}{lcccc}
        \toprule
        \textbf{Method} &
        \textbf{Color} &
        \textbf{Counting} &
        \textbf{Position} &
        \textbf{Style} \\
        \midrule
        CAA vector          & \textbf{96.20} & \textbf{68.75} & \textbf{78.60} & \textbf{82.71} \\
        Random direction    & 0.00 & 0.00 & 0.00 & 0.00 \\
        Unrelated direction & 0.00 & 0.00 & 0.00 & 0.00 \\
        \bottomrule
    \end{tabular}
\end{table}

As shown in Table~\ref{tab:additional_baselines}, the target CAA vectors achieve high SSR across all four semantic categories, whereas both the norm-matched random directions and the unrelated semantic direction yield an SSR of 0.0\%. These results show that successful Und$\rightarrow$Gen steering cannot be explained solely by the magnitude of the intervention or by injecting an arbitrary semantic direction. Instead, the steering effect depends on the target-specific semantic information encoded in the extracted vector.

\subsection{Additional Steering Vector Extraction Strategies Comparison in the Generation Branch}
\label{asec:object_centric_extraction} 

As described in Sec.~\ref{sec:gen_to_und} and Sec.~\ref{asec:gen_to_und}, we use global mean pooling over all VAE latent tokens as an architecture-agnostic strategy for extracting semantic steering vectors from the generation branch. This design avoids introducing additional external grounding modules, such as object detectors, segmentation masks, or attention-based token-selection heuristics, which may themselves introduce model- or category-specific biases.

However, global mean pooling may entangle information about object identity, background, spatial layout, illumination, global color statistics, and visual style. To examine whether this limitation accounts for the weak generation-to-understanding (\mbox{Gen$\rightarrow$Und}) transfer, we introduce two additional object-centric extraction strategies for ablation.

\begin{itemize}
    \item \textit{Object-aware pooling.}
We first localize the target object using its bounding box and map the localized image region to the corresponding VAE token positions. We then pool only the VAE tokens within this region, thereby excluding most background and scene-level information. Specifically, we obtain the bounding box using Grounding DINO~\cite{liu2023grounding}. For attribute concepts such as \texttt{red}, we use object-specific grounding prompts, such as \texttt{red book}, to improve localization accuracy.

 \item \textit{Attention-weighted pooling.}
We weight each VAE token according to its attention association with the target concept token, such as \texttt{red}, and compute a weighted average of the corresponding VAE hidden states. This strategy assigns greater weight to tokens that are more relevant to the target concept and reduces interference from background or semantically irrelevant tokens.
\end{itemize}

For each extraction strategy, we compute generation-branch steering vectors using CAA and apply them to the hidden states of the understanding branch during visual question answering. We evaluate the transfer using the steering success rate (SSR), which measures whether the intervention shifts the textual prediction toward the target semantic label. The results are reported in Table~\ref{tab:object_centric_gen_to_und}.

\begin{table}[t]
    \centering
    \small
    \caption{
    Gen$\rightarrow$Und steering success rate (\%) using different strategies for extracting semantic steering vectors from the generation branch.
    }
    \label{tab:object_centric_gen_to_und}
    \begin{tabular}{lcccc}
        \toprule
        \textbf{Generation-Branch Extraction} &
        \textbf{Color} &
        \textbf{Counting} &
        \textbf{Style} &
        \textbf{Appearance} \\
        \midrule
        Object-aware pooling      & 0.0 & 0.0 & 0.0 & 0.0 \\
        Attention-weighted pooling & 0.0 & 0.0 & 0.0 & 0.0 \\
        Global mean pooling        & 0.0 & 0.0 & 0.0 & 0.0 \\
        \bottomrule
    \end{tabular}
\end{table}

Across all three extraction strategies and four representative semantic categories, generation-derived vectors fail to produce successful Gen$\rightarrow$Und steering, with SSR remaining at 0.0\%. This result is also consistent across a broad range of intervention strengths. Therefore, the observed Gen$\rightarrow$Und failure is not specific to global mean pooling and persists when vector extraction is either restricted to or weighted toward object-relevant VAE tokens.

These results provide further evidence that, under the evaluated extraction and intervention protocols, generation-derived directions are less aligned with the semantic structure used by the understanding branch.

\subsection{Additional Layer-wise Logit Difference Analysis}
\label{asec:log_diff}
In Sec.~\ref{sec:gen_to_und}, to evaluate the effectiveness of steering when applied to the understanding branch, we adopt the logit-difference metric $m_{\mathrm{LD}}$ for the \texttt{red}$\leftrightarrow$\texttt{blue} semantic direction, which measures the model’s preference between a target answer and its contrastive alternative~\cite{logitdiff}.

\begin{figure}[h]
\centering
\includegraphics[width=0.85\linewidth]{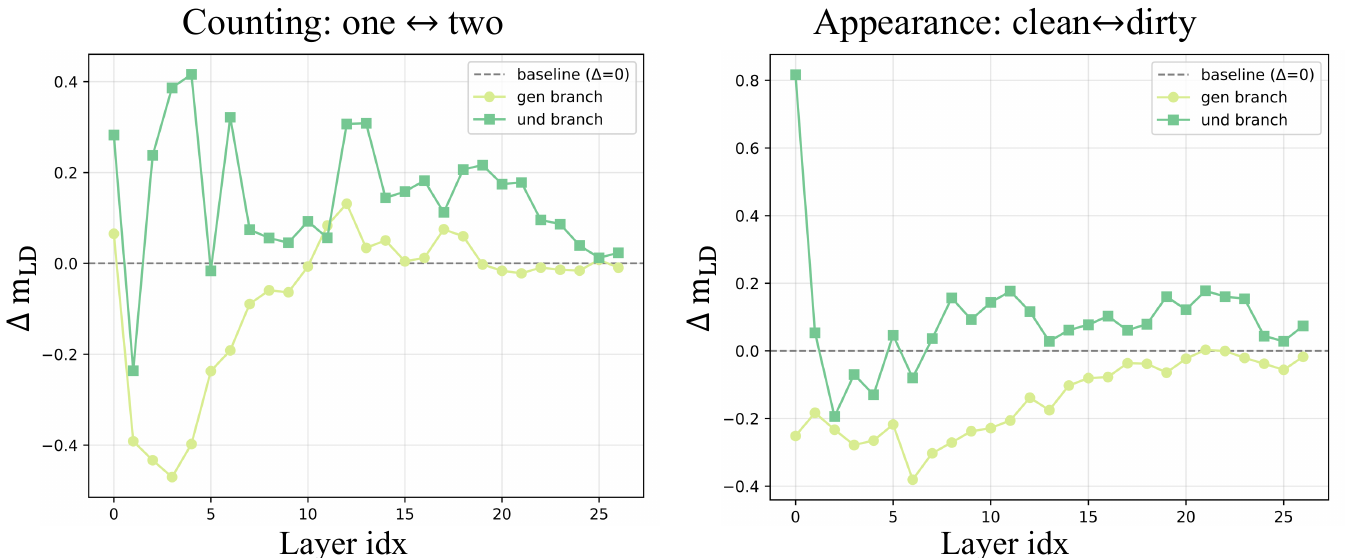}
\caption{Additional Layer-wise Logit Difference Analysis.}
\label{fig:more_logit_diff}
\end{figure}

Here, we provide additional layer-wise logit-difference results in Fig.~\ref{fig:more_logit_diff} for other semantic concepts, including counting (\textit{one}$\leftrightarrow$\textit{two}) and appearance (\textit{clean}$\leftrightarrow$\textit{dirty}).
Across these settings, steering vectors derived from the understanding branch consistently produce positive $\Delta m_{\mathrm{LD}}$ across layers, indicating a reliable shift toward the target semantic attribute. In contrast, steering vectors derived from the generation branch fail to induce meaningful changes in the model’s predictions.
Their corresponding $\Delta m_{\mathrm{LD}}$ values remain close to zero or even negative in early layers, suggesting that generation-derived directions are misaligned with the semantic representations used in the understanding branch.

These results are consistent with the main findings and further confirm that cross-branch transfer is asymmetric: only understanding-derived semantic directions can effectively influence the understanding process, while generation-derived directions lack the necessary alignment to modify the model's predictions.

\subsection{PCA Subspace Projection for Steering Analysis}
\label{asec:psp}
In Sec.~\ref{sec:why_asymmetry}, we observe a clear asymmetry: steering fails from generation to understanding, while the reverse direction succeeds. To analyze this phenomenon, we hypothesize that semantic directions learned in the generation branch are not well aligned with the semantic structure used by the understanding branch.

To quantitatively verify this hypothesis, we introduce the \emph{PCA Subspace Projection} (PSP) metric, which measures how well a steering vector aligns with a target semantic subspace induced by contrastive data. Intuitively, if a steering vector captures the intended semantic concept, it should lie within (or close to) the dominant subspace spanned by semantic variations of that concept.

\paragraph{PCA Subspace Projection (PSP) Formulation.}
First, we describe the computation pipeline of PSP as follows:
\textit{Step 1: Construct semantic difference vectors.}
Given a set of contrastive pairs, we extract the hidden representations corresponding to the target tokens (e.g., answer tokens) and compute difference vectors:
\begin{equation}
\mathbf{d}_i = \mathbf{h}_{+,i} - \mathbf{h}_{-,i},
\end{equation}
where $\mathbf{h}_{+,i}$ and $\mathbf{h}_{-,i}$ denote the hidden states of the positive and negative samples, respectively. These difference vectors isolate the semantic variation of interest (e.g., \textit{red} vs.\ \textit{blue}) while largely canceling out shared contextual information.

\textit{Step 2: Estimate the semantic subspace.}
We perform Principal Component Analysis (PCA) on the set $\{\mathbf{d}_i\}_{i=1}^N$ to capture the dominant directions of semantic variation. Let
\begin{equation}
U = [\mathbf{u}_1, \dots, \mathbf{u}_k] \in \mathbb{R}^{d \times k}
\end{equation}
denote the matrix of the top-$k$ principal components. In this paper, we set $k=20$ for all analysis studies. These orthonormal vectors span a low-dimensional subspace that approximates the intrinsic semantic structure of the target concept.

\textit{Step 3: Measure subspace alignment.}
Given a steering vector $\mathbf{s}$, we measure its alignment with the semantic subspace via the projection ratio:
\begin{equation}
\mathrm{PSP}(\mathbf{s}) = \frac{| U^\top \mathbf{s} |^2}{| \mathbf{s} |^2}.
\end{equation}
This quantity corresponds to the fraction of the energy of $\mathbf{s}$ that lies within the PCA subspace. A higher PSP value indicates stronger alignment with the semantic structure captured by the contrastive data, while a lower value suggests that the steering vector encodes directions orthogonal to the target semantics.

\begin{figure}[!t]
\centering
\includegraphics[width=0.85\linewidth]{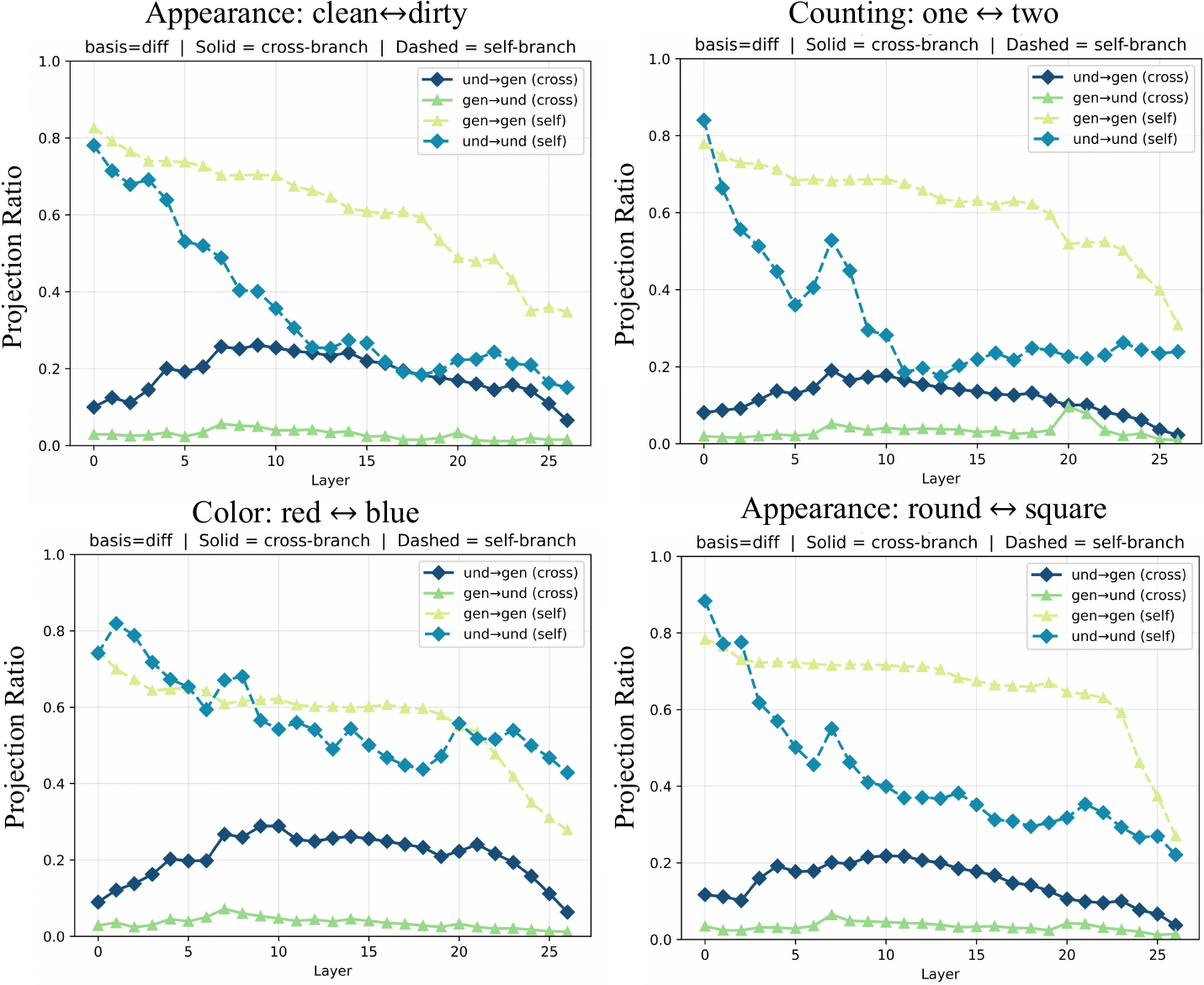}
\caption{Additional Layer-wise PCA Subspace Projection Analysis.}
\label{fig:more_pca_ana}
\end{figure}

\textbf{Additional PCA Subspace Projection Analysis.}
In Sec.~\ref{sec:psp}, we measure the PCA Subspace Projection (PSP) for the \texttt{red}$\leftrightarrow$\texttt{blue} semantic direction under both self-branch and cross-branch settings. Additional results across different concepts are shown in Fig.~\ref{fig:more_pca_ana}, exhibiting consistent patterns.
Steering vectors derived from the understanding branch show consistently high alignment in both settings, indicating that they capture generalizable and transferable semantic directions. In contrast, steering vectors from the generation branch exhibit low alignment when projected onto the understanding subspace, explaining their limited influence on downstream predictions in the understanding branch.
Overall, these results suggest that cross-branch transfer is governed by subspace alignment: only directions that lie within a unified semantic subspace can transfer effectively across branches.

\subsection{Generalization to Out-of-Template Compositional Prompts}
\label{asec:conceptmix}

Our primary evaluation uses controlled and largely template-based examples to isolate individual semantic factors and reduce potential confounding effects. This controlled setup enables a more reliable analysis of cross-branch steering, but does not fully reflect the complexity of free-form, real-world text-to-image prompts.

To evaluate whether the extracted understanding-derived steering directions generalize beyond this controlled setting, we conduct an additional out-of-template evaluation using ConceptMix~\cite{conceptmix}, a public compositional text-to-image benchmark containing free-form prompts that combine multiple visual concepts. We evaluate prompts at different compositional difficulty levels, denoted by $k$, where a larger $k$ indicates that more visual concepts are simultaneously specified in the prompt.

We consider two representative semantic categories, \textit{color} and \textit{counting}, and apply the corresponding understanding-derived steering vectors to the generation branch. We follow the same steering success rate (SSR) evaluation protocol as in the main experiments. The results are reported in Table~\ref{tab:conceptmix_steering}.

\begin{table}[t]
    \centering
    \small
    \caption{
    Und$\rightarrow$Gen steering success rate (SSR, \%) on out-of-template compositional prompts from ConceptMix. The compositional difficulty increases with $k$.
    }
    \label{tab:conceptmix_steering}
    \begin{tabular}{lccc}
        \toprule
        \textbf{Semantic Concept} &
        $\boldsymbol{k=2}$ &
        $\boldsymbol{k=3}$ &
        $\boldsymbol{k=4}$ \\
        \midrule
        Color    & 46.25 & 21.43 & 20.45 \\
        Counting & 42.22 & 33.33 & 17.28 \\
        \bottomrule
    \end{tabular}
\end{table}

As shown in Table~\ref{tab:conceptmix_steering}, the understanding-derived directions retain non-zero steering effectiveness on free-form prompts containing multiple simultaneous visual concepts. This indicates that the extracted semantic directions are not limited to the controlled prompt templates used in our primary evaluation and can generalize, to some extent, to out-of-template compositional prompts.

At the same time, SSR generally decreases as the compositional difficulty $k$ increases. This suggests that controlling a target semantic attribute becomes more challenging when the prompt contains a larger number of interacting visual constraints. Overall, these results demonstrate partial generalization of Und$\rightarrow$Gen steering beyond the controlled evaluation setting, while also highlighting highly compositional generation as an important remaining challenge.

\subsection{Ablation study}

\begin{figure}[!t]
\centering
\includegraphics[width=0.6\linewidth]{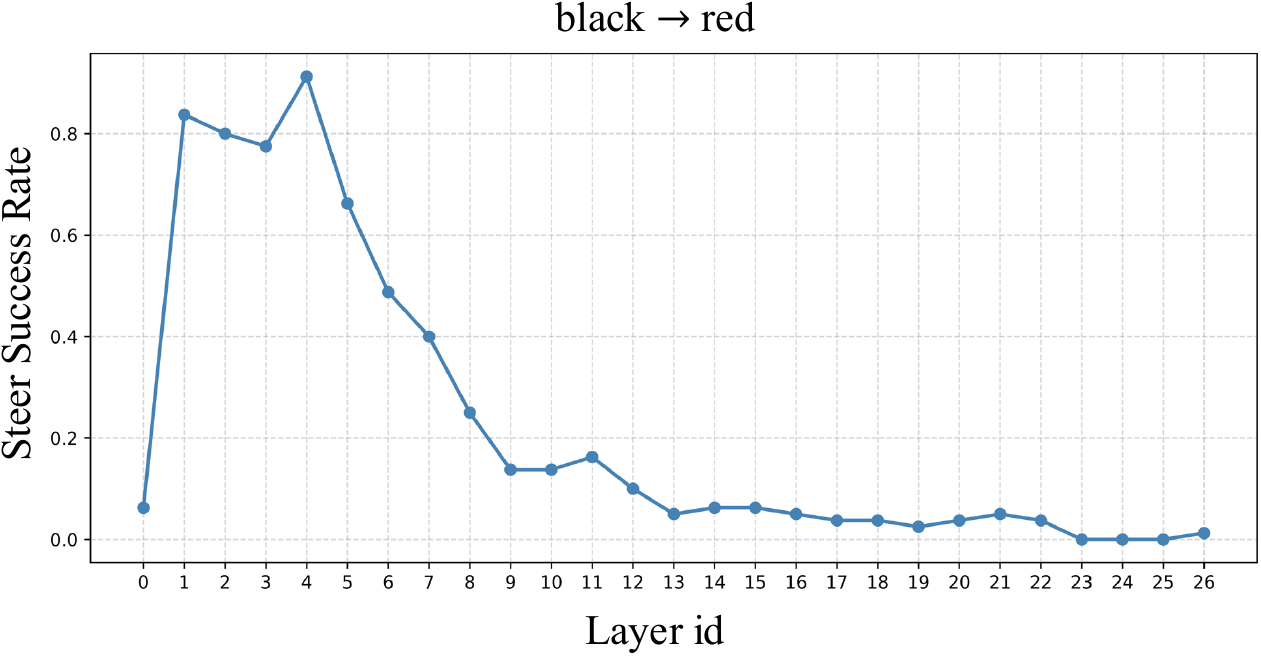}
\caption{\textbf{Ablation about the steering layer.} Steering is most effective in early-to-middle layers (2--5) and degrades in deeper layers, indicating that transferable semantics are primarily encoded in intermediate representations.}
\label{fig:layer_ablation}
\end{figure} 

\paragraph{Ablation study about the steering layer.}
In this work, we apply steering vectors to all layers by default. Here, we further analyze the effect of layer-wise steering by applying the steering vector to each layer individually. Specifically, we evaluate steering performance when the intervention is restricted to a single layer at a time, which allows us to examine the contribution of different layers to cross-branch transfer.

As shown in Fig.~\ref{fig:layer_ablation}, steering effectiveness varies significantly across layers. The performance peaks in early-to-middle layers (around layers 2--5), where steering achieves the highest steering success rates, indicating that these layers play a dominant role in encoding transferable semantic directions. In contrast, steering applied to deeper layers leads to a sharp performance drop, with success rates approaching zero in the final layers. This suggests that later layers are less responsive to external semantic interventions and may be more specialized for task-specific decoding. Overall, these results indicate that cross-branch semantic control is primarily mediated by intermediate representations rather than late-stage processing.

\paragraph{Ablation study about the steering strength $\alpha$.}
We study the effect of the steering strength $\alpha$ on cross-branch transfer performance using the \texttt{red} $\rightarrow$ \texttt{blue} concept. Specifically, we vary $\alpha \in \{0.1, 0.15, 0.2, 0.25, 0.3, 0.35, 0.4\}$ and evaluate steering effectiveness using the SSR metric in the generation branch.

As shown in Tab.~\ref{tab:alpha_ablation}, ITI~\cite{cls} demonstrates the strongest robustness across different $\alpha$ values, maintaining consistently high SSR, while RepE~\cite{pca} is the most sensitive to the choice of $\alpha$ and exhibits the weakest performance overall. CAA achieves the best performance at smaller $\alpha$, but degrades rapidly as $\alpha$ increases. 

Across all methods, we observe a consistent decline in SSR as $\alpha$ becomes larger, suggesting that excessive steering introduces strong perturbations that disrupt the underlying representations (i.e., over-steering), leading to degraded semantic control. This highlights a clear trade-off between insufficient steering (under-control) and excessive intervention.

Based on these observations, in all subsequent experiments across different semantic concepts, we set $\alpha = 0.1$ for CAA and $\alpha = 0.2$ for the other methods as default configurations.

\begin{table}[t]
\centering
\small
\begin{tabular}{c|ccccccc}
\toprule
$\alpha$ & 0.1 & 0.15 & 0.2 & 0.25 & 0.3 & 0.35 & 0.4 \\
\midrule
CAA~\cite{caa} & \textbf{100.0} & 96.25 & 91.25 & 81.25 & 48.75 & 23.75 & 10.0  \\
RepE~\cite{pca} & 1.25 & 16.25& \textbf{20.00} & 12.5 &7.5 & 3.75& 1.25\\
ITI~\cite{cls} & 7.5 & 96.5 & \textbf{97.5} &97.5 & 97.5 & 96.25 & 93.75\\
\bottomrule
\end{tabular}
\vspace{5pt}
\caption{\textbf{Ablation on steering strength $\alpha$ for CAA (`red $\rightarrow$ blue').} SSR is reported.}
\label{tab:alpha_ablation}
\end{table}

\begin{figure}[!h]
\centering
\includegraphics[width=\linewidth]{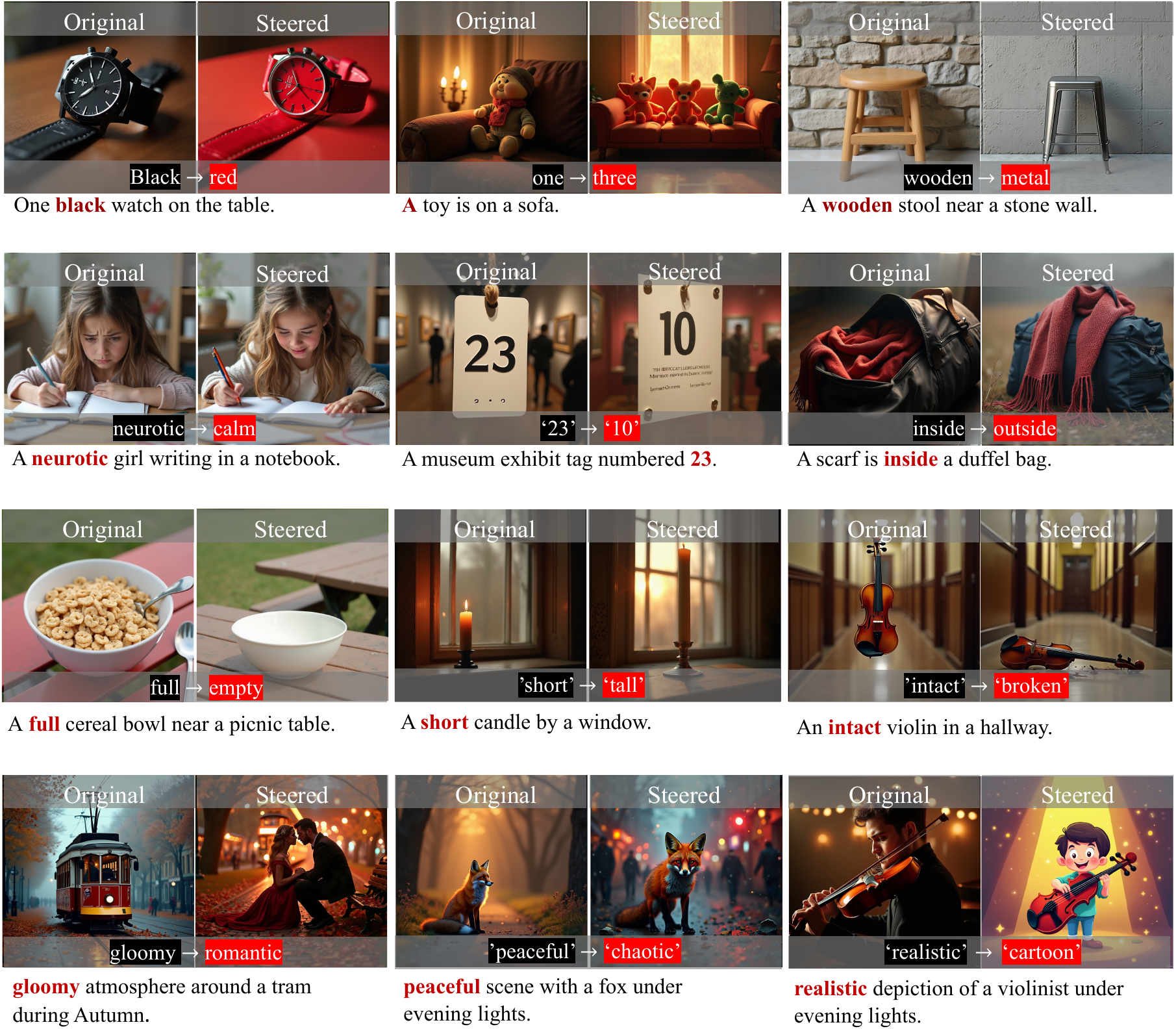}
\vspace{-10pt}
\caption{\textbf{Additional qualitative results of understanding$\leftrightarrow$generation steering.} Each pair shows the original generation (left) and the steered output (right). Steering modifies the target attribute while preserving object identity. }
\label{fig:more_vis}
\end{figure} 

\begin{figure}[!h]
\centering
\includegraphics[width=0.8\linewidth]{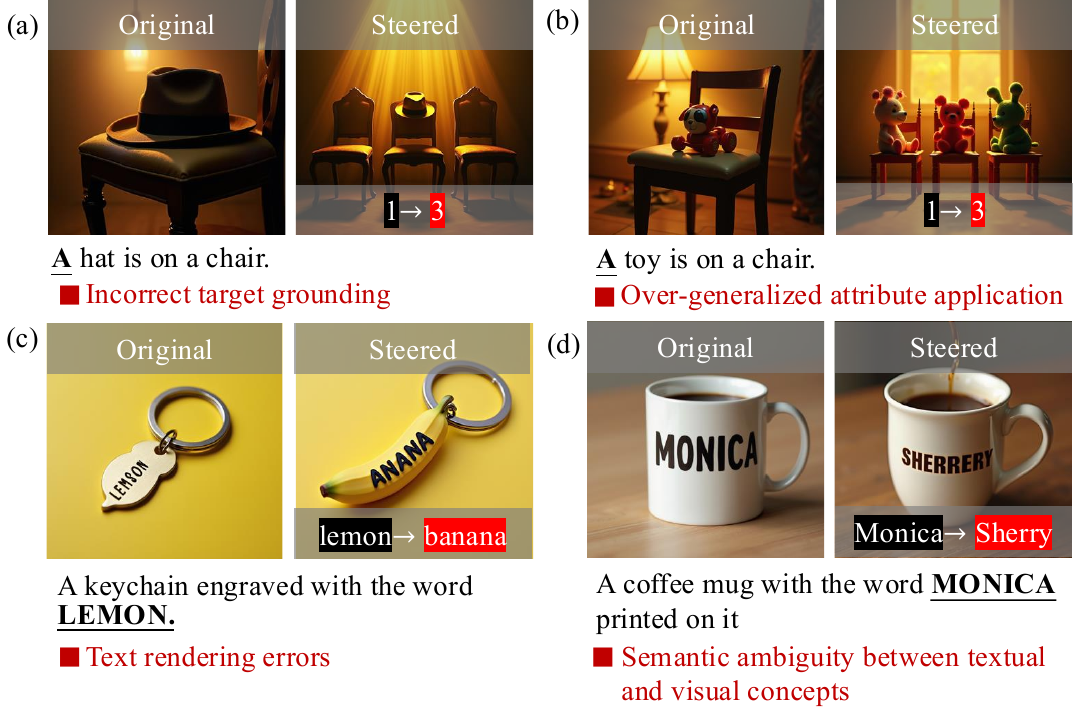}
\vspace{-5pt}
\caption{\textbf{Additional qualitative results of understanding$\leftrightarrow$generation steering.} Each pair shows the original generation (left) and the steered output (right). Steering modifies the target attribute while preserving object identity. }
\label{fig:failure_case}
\end{figure}

\subsection{Ablation Study on Gen$\rightarrow$Und Intervention Positions}
\label{asec:gen_to_und_intervention_positions}

In our original generation-to-understanding (Gen$\rightarrow$Und) steering setup, we add the generation-derived steering vector to the hidden states of all text tokens. This design is consistent with the understanding-to-generation (Und$\rightarrow$Gen) setting, in which the understanding-derived vector is applied to all VAE-token hidden states. It also avoids making an a priori assumption about which text-token position is primarily responsible for predicting the target semantic concept.

To examine whether the intervention target accounts for the weak Gen$\rightarrow$Und transfer, we additionally evaluate the following intervention positions:

\begin{itemize}
    \item \textit{All text tokens.}
    The steering vector is added to the hidden states of all text tokens, following our original setup.

    \item \textit{Concept-related input tokens.}
    The steering vector is applied only to input tokens associated with the queried object or attribute.

    \item \textit{Answer-prediction hidden state.}
    The steering vector is applied only to the final hidden state used to predict the answer token.
\end{itemize}

We evaluate these intervention positions using generation-derived vectors obtained with all three extraction strategies introduced in Sec.~\ref{asec:object_centric_extraction}: global mean pooling, object-aware pooling, and attention-weighted pooling. All other experimental settings, including the intervention layers and steering strengths, are kept unchanged.

Across all evaluated extraction strategies and intervention positions, the steering success rate remains 0.0\% for color, counting, style, and appearance. The result also remains unchanged when the steering strength is varied over a broad range. Therefore, the weak Gen$\rightarrow$Und transfer cannot be attributed solely to the use of all text-token hidden states as the intervention target.

These results show that the Gen$\rightarrow$Und failure persists across both extraction strategies and intervention positions. Under the evaluated protocols, generation-derived steering directions do not encode the target concepts in a form that can be directly transferred to and used to control the understanding branch.

\section{Additional Qualitative Result}
\subsection{Additional Qualitative Results for Understanding-to-Generation Steering}
\label{asec:visual_case}
In this section, we present additional qualitative examples in Fig.~\ref{fig:more_vis}, illustrating cross-branch steering from the understanding branch to the generation branch using CAA~\cite{caa}. Each pair consists of the original output (left) and the steered output (right).

Across all semantic categories in our \ourdata, steering consistently modifies the target attribute while preserving object identity and overall scene composition. For instance, color attributes are precisely altered (e.g., \textit{black} $\rightarrow$ \textit{red}, \textit{blue} $\rightarrow$ \textit{green}), object counts are adjusted without changing the scene layout (e.g., $1 \rightarrow 3$), and appearance and style are successfully manipulated (e.g., \textit{full} $\rightarrow$ \textit{empty}, \textit{gloomy} $\rightarrow$ \textit{romantic}). 

Overall, these qualitative results provide strong visual evidence that cross-branch steering from understanding to generation enables precise and compositional semantic control.

\subsection{Failure Case Analysis}
\label{asec:fase_case}
As shown in Fig.~\ref{fig:failure_case}, we further present representative failure cases of CAA-based understanding-to-generation steering.
Although CAA generally enables precise cross-branch control, failures still arise when the model cannot reliably localize the target object or disentangle the intended semantic attribute.

We observe four main failure modes:
(\textbf{i}) \emph{Incorrect target grounding}, as shown in Fig.~
\ref{fig:failure_case}(a), where the steering is applied to a correlated but unintended object;
(\textbf{ii}) \emph{Over-generalized attribute application}, see Fig.~
\ref{fig:failure_case}(b), where the semantic direction affects multiple objects rather than the intended target;
(\textbf{iii}) \emph{Text rendering errors}, where the intended textual modification is not faithfully realized, e.g., incorrect spelling in Fig.~
\ref{fig:failure_case}(c);
(\textbf{iv}) \emph{Semantic ambiguity between textual and visual concepts}, see Fig.~
\ref{fig:failure_case}(d), where the model confuses text content with visual object attributes, e.g., altering object shape instead of the engraved word.

These results suggest that remaining errors stem from limitations in object-level grounding, insufficient disentanglement of semantic directions, and challenges in faithfully controlling discrete textual outputs within the generation process.

\section{Societal Impact}
Our work has both potential positive and negative societal impacts. 
On the positive side, cross-branch semantic steering provides a diagnostic tool for analyzing whether unified multimodal models share transferable semantic representations across understanding and generation. 
This may contribute to more interpretable, controllable, and reliable multimodal generation systems, and can help identify representational mismatches that lead to semantic errors or hallucinations. 
On the negative side, techniques that improve controllability over generated content may also be misused to manipulate visual outputs, create misleading images, or amplify harmful biases if deployed without safeguards. 
Therefore, we view our framework primarily as an analysis and diagnostic tool, and emphasize the need for careful evaluation, transparency, and safeguards when applying steering-based interventions in real-world systems.

\section{Limitations}
\label{asec:limitation}
While our study provides evidence for asymmetric cross-branch transfer in UMMs, several limitations remain.
First, our analysis focuses on a predefined set of semantic concepts constructed from contrastive pairs. Although our additional evaluation on out-of-template compositional prompts demonstrates that the extracted steering directions can generalize beyond the controlled setting, steering effectiveness decreases as prompts contain more simultaneous visual concepts. This suggests that reliably controlling target semantics under complex attribute compositions remains challenging.
Moreover, our evaluation relies on automated judgment using a strong multimodal model. Although we perform human verification to ensure reliability, such evaluation may still introduce biases or miss subtle semantic variations.

\end{document}